\documentclass[conference]{IEEEtran}
\usepackage{graphicx}
\usepackage{verbatim }
\usepackage{xcolor}
\usepackage{caption}
\usepackage{subcaption}

\begin{document}

\title{Episodic Memory Model for Learning Robotic Manipulation Tasks}


\author{\IEEEauthorblockN{Sanaz B. Behbahani, Siddharth Chhatpar, Said Zahrai, Vishakh Duggal and Mohak Sukhwani}
\IEEEauthorblockA{ABB Discrete Automation and Robotics\\
sanaz.b.behbahani@us.abb.com, siddharth.chhatpar@us.abb.com, said.zahrai@de.abb.com,\\ vishakh.duggal@in.abb.com, mohak.sukhwani@in.abb.com
}}

\maketitle

\begin{abstract}
Machine learning, artificial intelligence and especially deep learning based approaches are often used to simplify or eliminate the burden of programming industrial robots. Using these approaches robots inherently learn a skill instead of being programmed using strict and tedious programming instructions. While deep learning is effective in making robots learn skills, it does not offer a practical route for teaching a complete task, such as assembly or machine tending, where a complex logic must be understood and related sub-tasks need to be performed. We present a model similar to an episodic memory that allows robots to comprehend sequences of actions using single demonstration and perform them properly and accurately. The algorithm identifies and recognizes the changes in the states of the system and memorizes how to execute the necessary tasks in order to make those changes. This allows the robot to decompose the tasks into smaller sub-tasks, retain the essential steps, and remember how they have been performed.
\end{abstract}

\IEEEpeerreviewmaketitle


\section{Introduction}

A typical automation system includes, in addition to the robot itself, various elements that collectively make the intended process feasible. Examples of such elements are grippers, fixtures, cameras, sensors, and simple passive elements such as trays and containers. The system composed of such elements needs to be programmed so that action sequences take place in an expected and controlled manner to achieve the desired goal. The conventional programming process for industrial robots includes representing all the elements that are present in the process and programming how and in what order the interactions between them should occur. Each program is case-specific and typically needs weeks or months of implementation and verification before the production starts~\cite{ref01, ref02}.

To enhance the re-usability of such systems, programs are usually divided into more diminutive and reusable modules or  routines, which can be utilized in other applications. To further simplify the work, one can provide a framework with all essential parts prepared and introduced~\cite{ref03, ref04}. As long as the template explicitly covers the case of interest, it can minimize the programming time considerably~\cite{ref05, ref06, ref07}. Graphical tools and visual programming languages are also often utilized to help the programming process~\cite{ref08, ref09, ref10, ref11}. 

All of the above mentioned methods can be recognized as standard techniques to reduce the burden of programming and make it more effective. Despite all the attempts to simplify the programming of robots, a single and comprehensive method that eliminates or reduces the programming effort for all kind of applications does not exist and unless templates are prepared for a specific domain, an expert programmer with knowledge from a specific robot is needed to develop an efficient and reliable robot application for industrial use.

Recent advances in machine learning, computer vision, and artificial intelligence, together with availability of affordable high-performance computing power have led to the initiation and development of new ideas for making robots more accessible to inexperienced and first time robot users. Innovative strategies are proposed and developed to make the robots trainable rather than programmable.

Learning from Demonstration (LfD) or Programming by Demonstration (PbD) denote methods, which have received wide attention from the robotic community. The fundamental idea is to enable the robot to learn and perform tasks without programming in a similar way as humans~\cite{ref12, ref13, ref14}. In a nutshell, a human tutor will demonstrate specific tasks to a robot, and the robot will use the demonstrated knowledge to execute the instructed task. The LfD process usually has an off-line learning phase, from which the data collected from successfully performed tasks are used for training. Typically, Gaussian Mixture Models (GMM), Bayesian networks, or dynamical system control are used to learn the trajectory or the action sequence. During operation, the input from a set of sensors is used to control the robot. As long as the input is consistent with the training cases, with some level of robustness to environmental noise and perturbations, the robot will behave properly. If the system faces an unknown condition, incompatible with the training data, the system does its best to perform, but the outcome will not be predictable~\cite{ref15, ref16, ref17, ref19}. In general, these methods require a large amount of data or prior knowledge for each new skill.

A more recent, forward-looking approach to the LfD, is the one-shot learning or learning from a single demonstration. In some approaches, the concept of learning from demonstration is joined with some hierarchical task networks and is accomplished via the bi-directional communication between the tutor and robot, where both are committed to the shared goal of the interaction and both actively contribute towards achieving a single goal~\cite{ref18}. More advanced strategies to one-shot learning combine meta-learning with imitation, enabling a robot to reuse previous experience and learn new skills from a single demonstration using deep learning methodologies~\cite{ref20}.

In deep learning (DL) approaches, high capacity, deep network models are used to generalize to the open world. The major downside to the DL approaches is that they require large amounts of diverse training data to be effective, therefore, it may not be suitable for novel robot users ~\cite{dasari2019robonet}.
Reinforcement learning (RL) is another novel method of robot learning. In this approach, the focus is on the manipulated object at any time, where the demonstrations are generated automatically through RL rather than the human demonstration. In this approach, the goal is to solve sequential decision-making problems by policy learning~\cite{ref23, ref24}.
Deep Q-networks are the most famous methods for deep reinforcement learning, where the action value function is approximated by a neural network and the value functions get updated by the Q-network~\cite{sasaki2017study}. 

\begin{figure}
 \centering
  \includegraphics[width=0.9\linewidth]{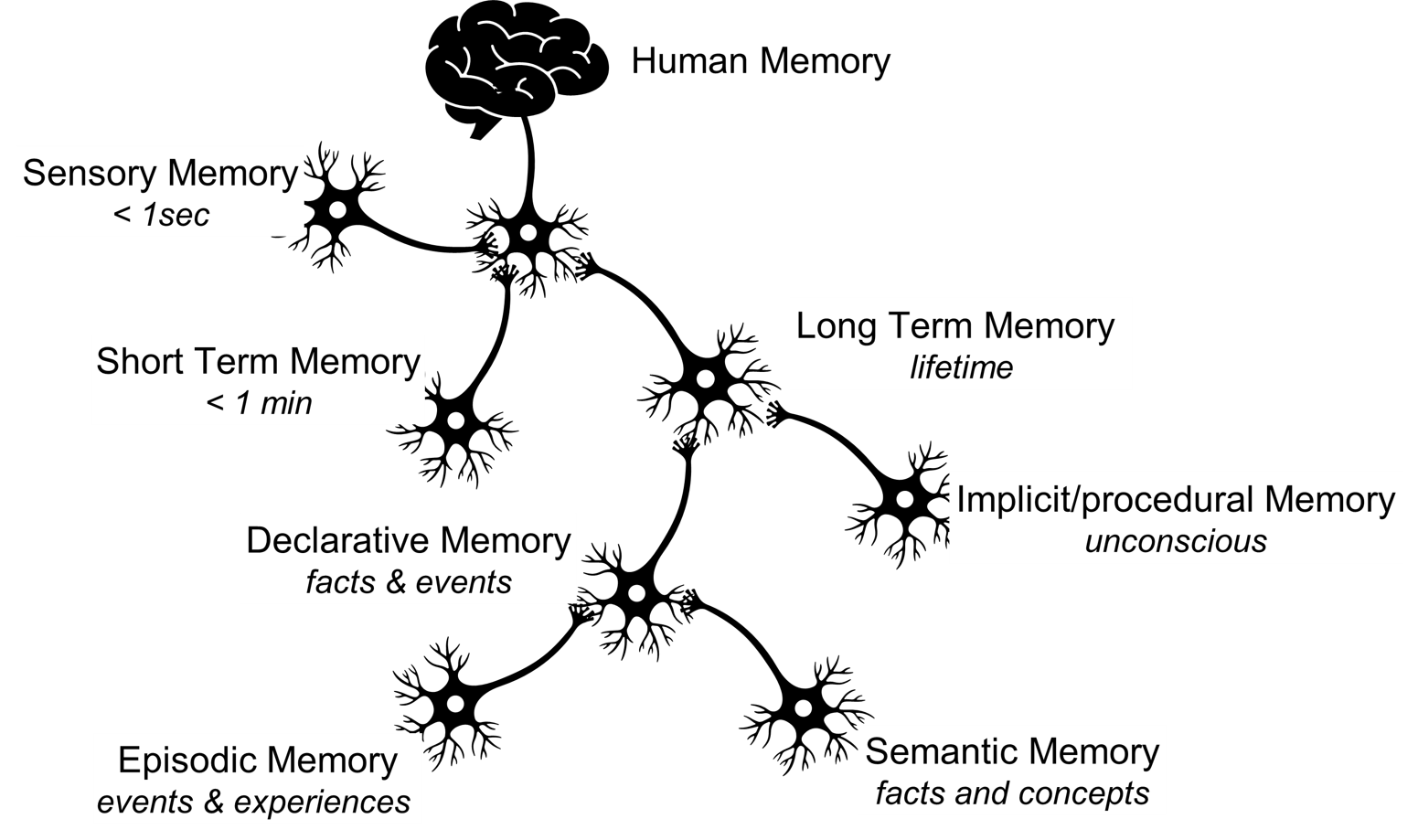}
  \caption{Structure of human memory.}
  \label{fig:HumanMemory}
\end{figure}

Learning is an ability possessed by humans, animals, and certain plants, and concerns acquiring understanding, knowledge, behaviors, skills, values, and preferences~\cite{wiki_learning}. Acquired knowledge and information are collected in the memory and prepared for later use. In general, memory can be short-term or long-term. As far as the learning concerns, long-term memory should be considered. According to neuroscientists, long-term memory can be divided into procedural and declarative. Procedural or implicit memory is the unconscious memory that is used for learning skills, such as balancing, biking, and playing musical instruments. Declarative memory is where facts and events are stored and can be recalled consciously. Declarative memory is further divided into episodic and semantic memory, as illustrated in figure~\ref{fig:HumanMemory}. Episodic memory stores knowledge about events, while semantic memory accumulates facts and data together with logical relations~\cite{h-mem}.

Procedural memory is typically acquired through repetition and practice and is composed of unconsciously relating sensory inputs to motoric behaviors. Artificial neural networks are used to model procedural memory and allow a fast recognition or reaction to sensory data after a complete training~\cite{ref26}.
Semantic memory is developed through observations, training, and consciously acquiring information. For example, we learn the laws of physics and apply them to solve problems and perform our daily tasks. Semantic memory can for example be modeled by expert systems~\cite{ref25}. In~\cite{ref27} a knowledge graph was used as the core of the semantic memory. 
It enables the robot to learn a new task, by shadowing a tutor, learning, and generalizing from only one demonstration.
Episodic memory is a long-term memory of experiences and memories of events. Episodic memory provides means to store various experiences and combine them to a more abstract construct and recall consciously when needed~\cite{ref29, ref30, ref31}. 

While all of above methods are contributing to ease of working with robots, none of them have reached a degree of reliability and generality to enter manufacturing industry, where the expectations on predictability, reliability and accuracy of outcomes are extremely high. In this work, our goal is to build a model for episodic memory that enables the user to teach the robot to perform necessary instructions in an industrial manufacturing application through one single demonstration, the same way as teaching an apprentice. 
Unlike typical LfD methods, our single shot teaching provides an economic, efficient, and convenient way of providing the robot with the required information to complete the desired task without programming~\cite{ref21, ref22}.

\section{Related Work}

Industrial robots have evolved from being pure engineering systems to become collaborative over time. Collaborative robots are designed to augment humans' abilities and act as an extension in work in complex industrial environments. To enhance the interactions, collaborative robotic architectures attempt to recreate human cognition and intend to reach a higher degree of autonomy by hierarchic control system design, environment sensing, and information fusion~\cite{ref27, mandal}. 

Recent researches have successfully demonstrated task-based dialogues with collaborative robots~\cite{ref27}. The proposed cognitive architecture can acquire a new understanding of objects, their properties, and details of the (manipulative) actions performed on the object. The authors have defined a graph-based memory design that sets the methodology apart from present-day data-hungry statistics-based approaches. The described architecture is robust for knowledge sharing among multiple industrial robots. This system allows users to teach the robot new knowledge about objects, activities, and tasks in one shot~\cite{thining}. One can transfer the learning and experiences from one robot to another in a simple manner. To be optimally useful and easy to use, robots need to effectively remember repetitive manipulative experiences. Episodic memory has an effective structure to store explicit past experiences and plays an essential role in diverse cognitive processes. It helps to put knowledge in order and structure it for future use. 

Various strategies for managing computational models of episodic memory are available to memorize experience sequences~\cite{rinkus2004neural, nuxoll2012enhancing, starzyk2009spatio}. Hierarchical emotional episodic memory, using deep adaptive resonance theory network~\cite{lee2018hierarchical} has also been proposed to learn emotions correlated with past experiences. Generalized fusion adaptive resonance theory-based EM-ART has been introduced to make the episodic memory suitable to handle intricate relations of past events~\cite{wang2012neural, subagdja2015neural}. Temporal event sequences are efficiently learned and recalled from stored episodes. Such elements are required to make intelligent robots for drastically simpler automation. 

The proposed method presented in this paper considers remembering essential events in the right order from software architecture and information management perspective. The method enables the robot to learn a new task, by shadowing a tutor, without the need for programming knowledge. We define a unified approach using graph-based finite-state automata. The behavior of the elements used in the application is modeled by state machines. Each element and its state machine can be utilized in any other application without the necessity for adaptation. A software architecture is introduced that simplifies building large robotic systems.

\section{Learning work instructions}
\subsection{Apprentice at work}
In the approach we present here, we aim to provide an industrial robot the ability to memorize the steps in a process and repeat them when requested, similar to what an apprentice would learn working in an industrial environment. Before going into the details of the methodology, it is worth imagining a learning scenario for our apprentice on the very first working day.

On the first day of work, the apprentice is introduced to the working environment. Tools and procedures are described, and the apprentice observes how each device works and if needed, receives necessary training. Next, the instructions are described by the tutor. The apprentice observes what is happening and how objects change in the working space. If our apprentice has a perfect memory, all steps will be memorized and repeated in the right order when needed. Our apprentice learns by observing changes in the environment and processes the information, but what is learned will depend on personal judgments and prior experiences as well as the attention paid to the details of the process. An apprentice that does not pay significant attention will not be able to memorize the steps properly and if there is a lack of experience to understand the observations, the learning process will not be complete. As an example, an adolescent will have complications learning to perform a complex assembly task, even if mechanical capabilities are sufficient.

To give our robots the ability to learn instructions and reproduce them, it must first be able to make observations on the objects it is working with. Secondly, the robot needs to understand the variations in the object and the surrounding world, record and retrieve both the changes and the process that has caused them. We accomplish this by decomposing the world around the robot into a collection of elements and describe their behavior, as well as interactions between them, in an observable and measurable manner. Details of the proposed system are described in the next section through an example of machine tending.

\subsection{Machine tending application}
A typical machine tending application has the following workflow. There is an infeed tray that contains parts that should be placed in the machine. The task starts by loading the machine with parts from this tray. After safely loading the machine, the door is closed and the machine is instructed to start the process. Once the process is complete, the machine stops, the door opens and the system is ready to remove the part from the machine and put it in the outfeed tray. In our example, we assume that the machine performs a test with a result that could be good or bad. Depending on the result, the parts are sorted in the respective trays.

\begin{figure}
  \centering
  \includegraphics[width=0.9\linewidth]{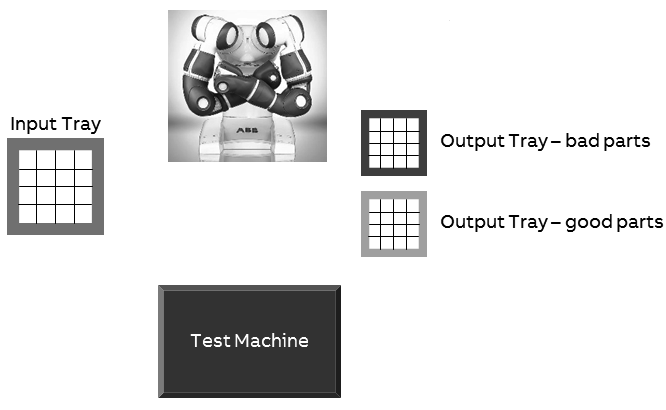}
  \caption{A simplified machine tending application work-cell.}
  \label{fig:workcell}
\end{figure}

Learning this process for an apprentice will not be challenging and it will be likely adequate to see the process only once. The apprentice will understand that the object should be picked from the in-feed tray and be placed in the machine, or be picked from the machine and placed in the proper outfeed trays, depending on the result of the test. Not picking from empty slots and not placing in full slots are explicit rules for an apprentice. We intend to build a robotic system that is capable of learning the process similarly. We use the case of the testing machine as an example, but the method is by no means restricted to any particular applications.

\begin{figure}[b]
  \includegraphics[width=\linewidth]{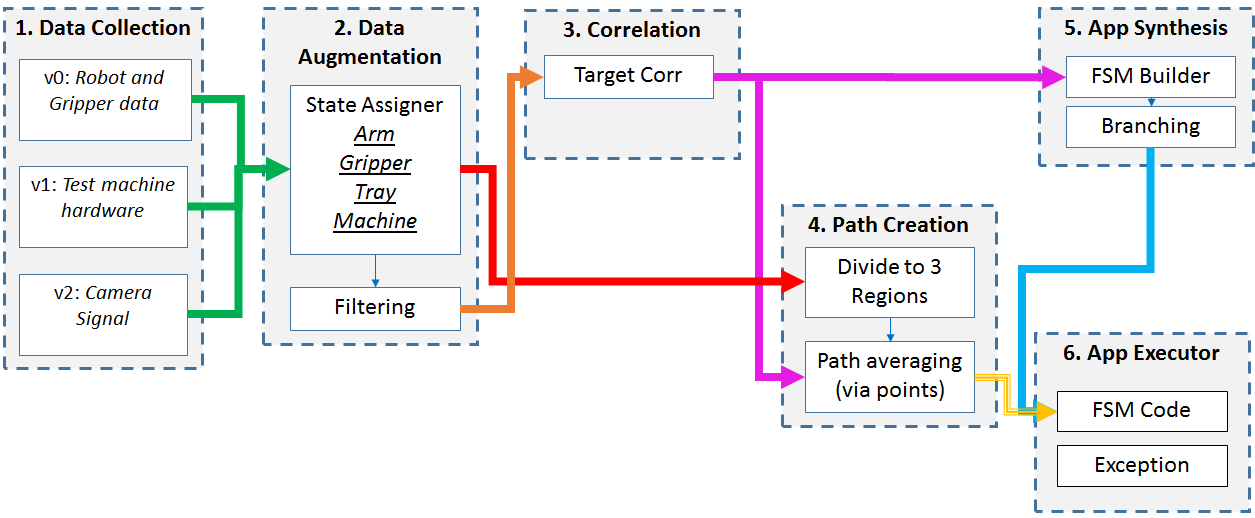}
  \caption{Architecture for learning and executing a Machine Tending Application.}
  \label{fig:architecture}
\end{figure}

\subsection{Solution}
A simplistic structure of a machine tending cell is shown in Figure \ref{fig:workcell} and consists of a robotic arm, grippers, trays, and a test machine. Although it is conventional to use machine vision to recognize and localize objects, here, for simplicity, we overlook that and instead assume that the objects are known and the locations are defined. Each component of this system is represented by a finite state machine, which closely describes the characteristic behavior of that element, regardless of the application it is being used in. Note that in a production system, the objects in the work cell are recognized by the vision system, their position and their configurations will be distinguished automatically.

Figure \ref{fig:architecture} shows the construction of the system for observing, learning, and executing the application. First, the data concerning all observable elements is collected and analyzed. The data can be augmented to advance the understanding and categorizing ability of the system. The data is first used to generate the path for moving elements of the system. Correlations between signals are used for discovering the essential events that change the state of one or several elements in the system.

How the environment is changing is a combination of the changes in the elements of the system. After the correlations are found and the state of each element is identified, the information is integrated into a transition to a new state of the environment, with the conditions required for the transition to happen. If the new state is the same as the ones previously visited, a closed loop is created and can be executed repeatedly. The series of state transitions form the application, for which a code can be generated and executed later. The proposed solution and the procedure of learning is described as follows:

\begin{itemize}
    \item A scene is separated into elements whose interactions lead to the execution of the desired task. These elements are identified either with a vision system or the information is fed to the system during commissioning as configuration. 
    \item A state machine is connected to each element, which includes the states of the element and the conditions that would lead to a transition from one state to another. These state machines are not application dependent but are constructed assuming that the element is used in a robotic system.
    \item All the existing elements are attached to the system and their data will become observable for the data collector.
    \item The robot and tools connected to it are the players in the system. The demonstration concerns leading the players so that the task can be accomplished. During the demonstration, the following steps are taken:
    \begin{enumerate}
        \item The data collector accumulates all data at each time step.
        \item The data is filtered and correlations, as well as the state changes for each element, are found.
        \item The state changes, as well as conditions leading to them, are synthesized in a state machine describing the changes in the scene.
        \item The recorded data is used to define the path for movable elements, like the robot itself.
        \item A loop is identified if one of the previous states is revisited.
        \item The synthesized state machine, together with the corresponding path, are saved as a program.
    \end{enumerate}
    \item The recorded program can be executed on command.
\end{itemize}
 
With the above method, our robot learns the steps necessary to perform a task, very similar to an apprentice who pays full attention during the demonstration and memorizes all details. In the next section, we present the details of implementation for our specific case of machine tending with an ABB robot.

\section{Implementation}
In this section, we explain the system developed for machine tending with ABB’s industrial robot IRB 14000, YuMi. YuMi is a two-armed robot with integrated grippers and lead-through functionality that allows the demonstration of tasks to be convenient. The tutor simply moves the robot arm in lead-through mode and demonstrates the desired task. During this operation,  data are collected by the data collection module, and the raw data are analyzed by the data augmentation module to extract element states. 

\begin{table*}[t]
\centering\begin{tabular}{ |l|c|c|c|c|c|c|c|c|c| }
\hline
~\textbf{Time}&\( t_0 \)  & \( t_1 \)  & \( t_2 \)  & \( t_3 \)  & \( t_4\)  & \( t_5 \)  & \( t_6 \)  & \(t_7\)  & \( t_8 \) \\
\hline
~\textbf{Robot position}&\( SS_0 \)  & \( SS_0 \)  & \( SS_0 \)  & \( SS_0 \)  & \( RM_0\)& \( RM_0 \)  & \( RM_0 \)  & \( RM_0 \)  & \( SS_1 \) \\
\hline
~\textbf{Command}&\( MoveL \)  & \( - \)  & \( - \)  & \( - \)  & \( -\)  & \( - \)  & \( - \)  & \( - \)  & \( Grip \)\\
\hline
\end{tabular}
\vspace{0.5cm}
\caption{A sample state trajectory.}
\label{table:ta}
\end{table*}

This state vector is the basis for extracting the logic of the process. In the case of machine tending, the elements whose actions lead to changes in the environment are the robot and the gripper. Therefore, we pay special attention to the following system states: robot motion (RM), indicating the robot is moving, gripper motion (GM), occurring when the gripper is in motion, and stand still (SS), which is when the robot and gripper are not moving. A sample state trajectory is presented in Table~\ref{table:ta}. At each instance of time, the elements of the system are in a specific state. We monitor the element states and record when a state transition occurs.

Three properties are used to identify a state:
\begin{itemize}
    \item Type of state (e.g., Robot Motion, Gripper Motion, Standstill)
    \item Condition for transitioning into state
    \item Commands given in the state
\end{itemize}
Using the state identity, we can recognize when a state is repeated and whether there is a loop in the system. The task application program is being generated on-the-fly, which enables the robot to perform the same task without the need for programming and based on what the system has observed from the tutor. Machine tending is only an example of what the system is capable of doing. This learning framework can handle different variations, for example, several trays and test machines, process task sequence, etc without any changes.

\subsection{State machines}
The element state machines encode essential information about the operation of the individual elements and how they interact with each other. This is equivalent to the mental models built by the apprentice for the elements involved in the process s/he is observing. The information encoded in an element state machine includes the discrete physical and operational states that the element can be in, the conditions that lead it to transition between states, the labeling of certain states as {\it active} to indicate interaction with other elements, and the commands that can be given to the element. 

\begin{figure*}
     \centering
     \begin{subfigure}[b]{0.45\textwidth}
         \centering
         \includegraphics[width=\textwidth]{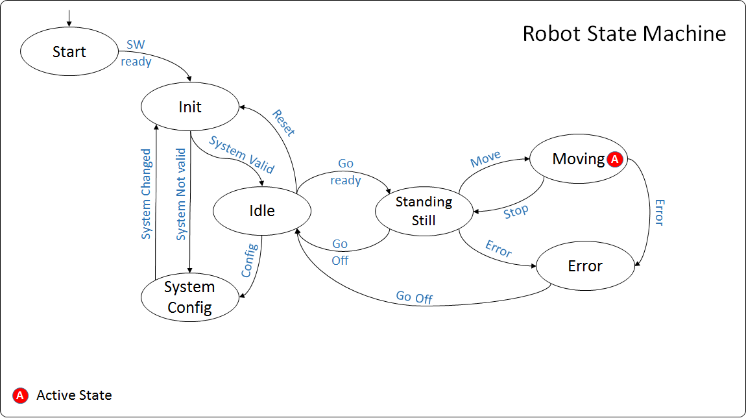}
         \caption{State machine describing the behavior of the robot.}
         \label{fig:robotsm}
     \end{subfigure}
     \hfill
     \begin{subfigure}[b]{0.45\textwidth}
         \centering
         \includegraphics[width=\textwidth]{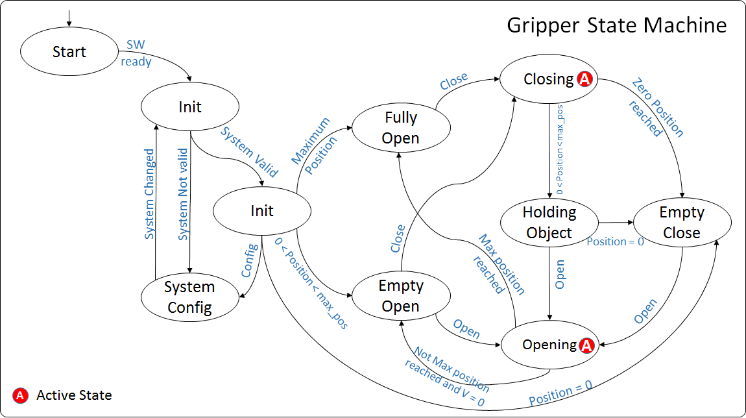}
         \caption{State machine describing the behavior of the gripper.}
         \label{fig:grippersm}
     \end{subfigure}
     \hfill
     \begin{subfigure}[b]{0.45\textwidth}
         \centering
         \includegraphics[width=\textwidth]{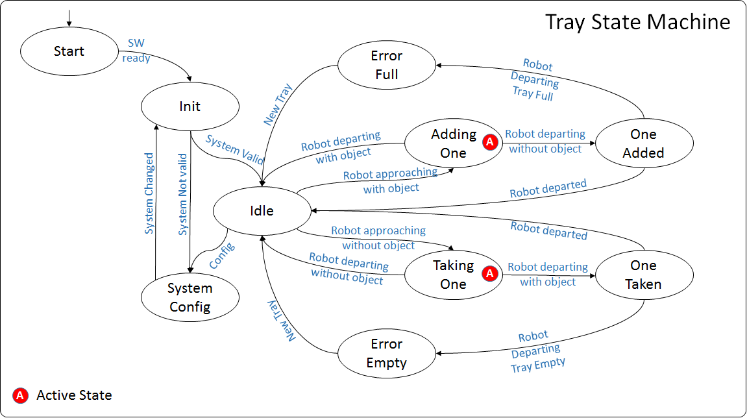}
         \caption{State machine describing the behavior of the tray.}
         \label{fig:traysm}
     \end{subfigure}
     \hfill
     \begin{subfigure}[b]{0.45\textwidth}
         \centering
         \includegraphics[width=\textwidth]{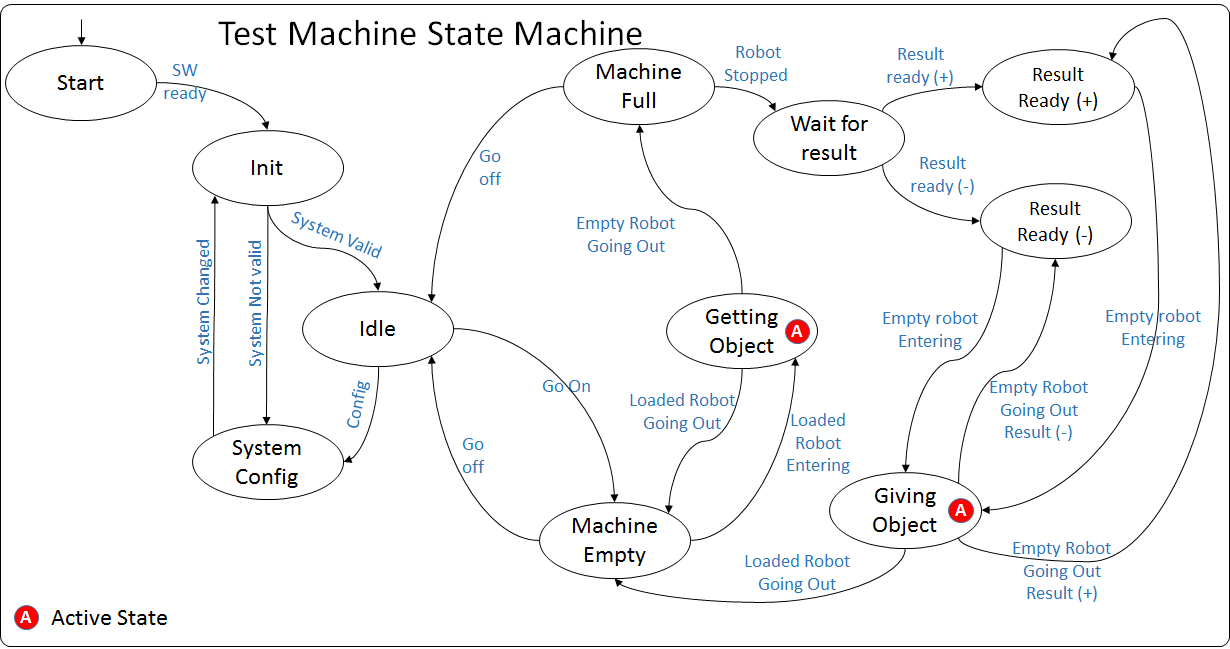}
         \caption{State machine describing the behavior of the test machine.}
         \label{fig:testmcsm}
     \end{subfigure}
        \caption{State machine description for different elements of the system.}
        \label{fig:sm}
\end{figure*}

Consider the example of the Test Machine State Machine (TMSM) (Figure ~\ref{fig:testmcsm}). It includes operational states that are common to all element state machines: {\it Start}, {\it Init}, {\it System Config}, and {\it Idle}. {\it Machine Full} and {\it Machine Empty} represent physical states where an object has been placed in the machine and when there is no object in the machine, respectively. A physical property common to both these states is that the machine door is closed. When the door is opened in the {\it Machine Empty} state, the state machine makes the transition to the {\it Ready to Get} state, where an object can be placed into the machine. 
The door opening is achieved via an “open door” command sent to the Test Machine Element. The TMSM encodes this information into the state machine for the benefit of the learning system by issuing an output with the command name when there is a transition from {\it Machine Empty} to {\it Ready to Get}. In general, every state transition has an output; this output is set to “None” if there is no command triggering that state change. 
In the {\it Ready to Get} state, the test machine is empty with its door open and ready to receive a test object. When a loaded robot enters the test machine in this state, the state transitions to a special state, called an “active state”, indicating that the element is in interaction with another element. There are two active states in the TMSM: {\it Getting Object} and {\it Giving Object}. The tagging of states as active allows the learning system to identify the interaction between elements. 
Other appropriate information can be included in the element’s state variables, e.g., the result generated by the test machine. This causes the {\it Test Running} state to be split into {\it Result Positive} and {\it Result Negative}, depending on the result generated, enabling the learning system to utilize this information to branch process control based on the result without any special means for encoding the decision making. 

Once the element state machines have been constructed in this manner, it is easy to see that a superstate machine can capture all the relevant information for the discrete process. The superstate machine aggregates the element states and has states representing seen element state combinations. Any information that is not captured in this process is a result of an insufficient design of one or more of the incorporated element state machines. Any state/event that is important to the process for selecting an “action” needs to be captured in an overall state change of the system so that the learning machine can recognize the need for taking action.

\subsection{Active elements}
In a robotic process, robot motions are almost always targeted towards interaction with other elements of the process, e.g., pick or put objects from/to tray, pick/put objects from/into test machine. Identifying and capturing these interactions are essential for learning the process. Also, the robot motions between interactions need to be understood as a single action with the target of getting into position to interact; meaning that the exact trajectory/motion profile is not important, but getting into position to interact is. To highlight this interaction, the concept of active elements is introduced. Elements are considered active when they enter into states that are labeled as active states. For example, the test machine is in interaction with the robot when the robot enters the machine to take or leave an object; therefore, the states {\it Getting Object} and {\it Giving Object} are active states in the TMSM. 

\subsection{Single shot learning of task}
During learning, the system builds an application state machine (ASM), which can later execute the learned process. The ASM build process is started with the inclusion of states that are common to all element state machines: {\it Start}, {\it Init}, {\it System Config}, and {\it Idle}, as well as an {\it Exception} state. As the process is demonstrated, the learning system adds states to the ASM. Because our implementation targets a robotic application, the types of states in our ASM were selected to be Robot Motion, Gripper Motion and Standstill; however, these can be arbitrarily chosen for different implementations. In fact, in a generic implementation new states can be added to the ASM whenever any of the underlying elements changes state without any correspondence to the state of the robot or gripper. 
With each new state, two transitions are added: one from the previous state to the new state and a second from the new state to the {\it Exception} state. 

The learning system also captures any commands produced by the element state machines and adds them to the appropriate state in the ASM. Commands can take arguments, which can be simple, e.g., the binary open/close commands for the gripper; or complex, e.g., location, in the case of robot motion command. Location information for robot motion commands is recorded in semantic form by the ASM, e.g., \{“Move to”, “Tray”\}. The semantic location is inferred by the correlation module based on what element is active when a robot completes its move. During execution, the semantic location is replaced by the known metric location for the Tray. 

In this manner, the learning system builds an initial application state machine as a string of states with each state having a transition to the next state in the process as well as the {\it Exception} state. This state machine is later filtered to detect loops and branches, resulting in a more complex state machine. 

When executing, the ASM transitions along the state chain as the underlying elements change state in accordance with the observed process. If a state change occurs that does not follow this process, i.e., the ASM sees a state combination that is not in its current state or the immediate next states that it can transition to, it transitions to the {\it Exception} state. The {\it Exception} state is a special state where the system interacts with a user describing why it is in this state and looks for direction on how to resolve the situation. The user can direct the system to {\it continue}, which is interpreted by the system to mean that the element state combination seen is an {\it allowed} combination in the state it was in. Or, if the user decides that special steps need to be taken to resolve the exceptional condition, s/he can request the system to {\it learn} and then proceed to demonstrate what to do in that condition. In this case, the system goes back to learning and adds new states to the ASM based on the user demonstration.

\subsection{Task state pruning}
The ASM is constructed as a linear chain of states. After the demonstration is complete, this linear chain is examined to detect any duplicate states. States are compared for duplexity based on their properties: type of state (Robot motion, Gripper motion, Standstill); transition condition (element state vector); any commands given in the state. If all the properties match for two states, then they are considered duplicates. In this case, the state that comes later in the chain is removed and the transitions to and from that state are instead pointed to the state that comes earlier in the chain. This is how loops get formed in the ASM. Loops can be genuine or erroneous. For example, when the robot is being moved to interface with a tray during a demonstration, jitter in successive position measurements at the boundary can indicate robot moving into and out of interfacing with the tray. This will result in duplicate states and loop formation. This loop is identified as an erroneous loop and discarded. Genuine loops can form if a process demonstration is repeated to show a variation. In case of the machine tending example, the user will demonstrate the process for the two possible results of the test. In this case, all the states leading up to the test machine producing a result are duplicates for the repeated demonstration. Therefore, during pruning, the duplicate states get replaced resulting in a branch formation in the ASM where the two branches represent the test machine producing a positive and a negative result, resp. 

\subsection{Execution of learned task}

Once an application is learned, the system stores it in its library. Any time the system is on, it is continuously monitoring the elements in its workcell. Based on the identified elements, it is able to load all the applications possible with those elements. When a particular application is selected, the system executes the corresponding ASM. The ASM moves to {\it Init} state and waits on a trigger -- in the case of the machine tending example: introduction of a full input tray. When a full input tray is introduced, the change in the tray element results in the ASM transitioning to a state where a command is given to the robot to move into interaction with the tray. As the robot starts moving, the change in the robot state machine results in the ASM transitioning to the corresponding robot motion state. In this manner, as long as the combined element state transitions follow the demonstrated process, the ASM executes normally. If a previously unseen condition is encountered, the ASM enters {\it Exception} state. In this state, the ASM interacts with the user to determine how to proceed.

\section{Software Platform}

\begin{figure}
  \centering
  \includegraphics[width=\linewidth]{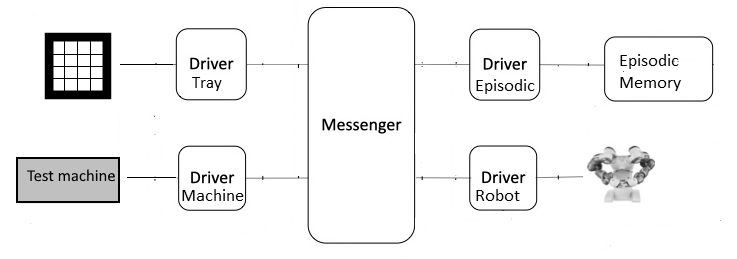}
  \caption{Platform architecture}
  \label{fig:platform}
\end{figure}

Figure ~\ref{fig:platform} presents the software architecture used for this implementation. The core of the platform is a message broker, RabbitMQ, to allow a flexible communication between different components. Any component in the system which either generates or consumes data, e.g. robot, tray, test machine, needs to be connected to the message broker. Messages are in JSON format to ensure cross language compatibility. Each component is wrapped by a driver, following a standard template. In what follows, we describe the robot driver, as an example that illustrates the structure of a driver and especially explain how the communication with the robot is implemented.

All drivers consist of three major sub modules: Façade, Decorator and Bridge. Façade acts as the interface between the driver and the message broker and is responsible to interpret JSON messages. Decorator implement the logic and the desired representation of the element. Bridge provides the communication link to the element. 

In the case of the robot, two arms and their respective grippers are packed in one driver, which communicates with the robot controller, running a hard real time application. The representation should be such that they can be considered as four different components, but can be operated synchronously if needed. Consequently, the driver needs to provide following functionalities: 
\begin{enumerate}
	\item Ability to control robot arms and their respective grippers independently.
	\item Lead through programming support.
	\item Access to data generated by robot in real time. 
	\item Synchronous execution of commands.
\end{enumerate}

To ensure that the system responds properly to the robot, the bridge is divided into 8 modules, each of which running in a separate thread, as follows:
\begin{itemize}
	\item Thread 1 and 2 execute motion commands for left and right arms of YuMi.
	\item Thread 3 and 4 execute gripper commands for left and right grippers.
	\item Thread 5 and 6 provide access to real time data generated by robot.
	\item Thread 7 and 8 allow stopping/re-starting of commands in execution.
\end{itemize}

The robot provides a TCP/IP server, implemented in ABB's programming language, RAPID, and the platform is a client consuming services offered by the server. 

\section{Experiment}
\begin{figure}
  \centering
  \includegraphics[width=0.8\linewidth]{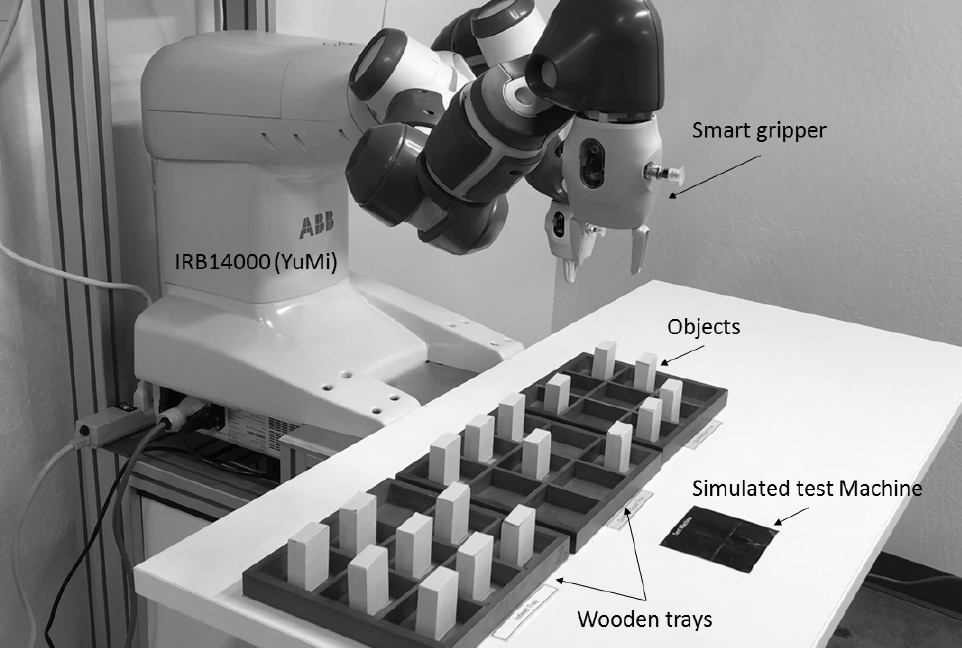}
  \caption{Experimental setup for machine tending.}
  \label{fig:model_machine_tending}
\end{figure}

The method described above was examined by applying to the case of machine tending explained earlier. The experiment was performed with the dual arm, collaborative IRB14000 (YuMi), equipped with smart gripper from ABB, with simple wooden blocks and trays as shown in Figure \ref{fig:model_machine_tending}. The test machine is simulated with a given location on the table, the black square, and the results was randomly set to resemble a realistic scenario.
 
When the system starts for the first time, it is expected to scan its environment and collect information about the environment. For the object in surrounding, this will typically mean that a vision sensor scans the scene and identifies the objects present in the scene. The system will also scan the interfaces, such as field busses, Ethernet ports, USB ports etc. to identify devices that are available. Although this is a standard procedure, in our test, we explicitly provide the information to simplify the implementation.

\begin{figure*}[t]
     \centering
     \begin{subfigure}[t]{0.3\textwidth}
         \centering
         \includegraphics[ width=\textwidth]{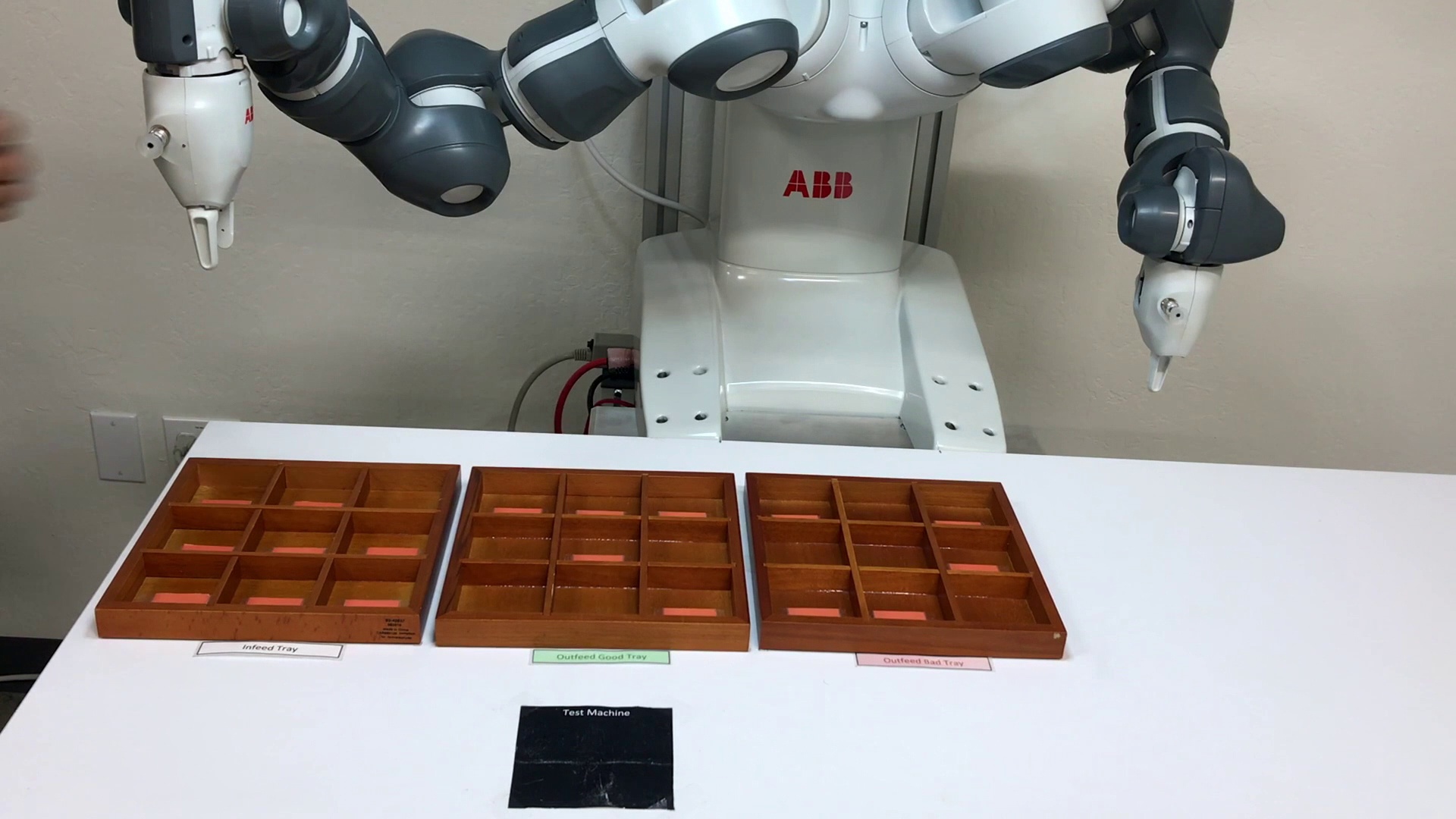}
         \caption{Initial state observed and understood by the system.}
         \label{fig:training01}
     \end{subfigure}
     \hfill
     \begin{subfigure}[t]{0.3\textwidth}
         \centering
         \includegraphics[width=\textwidth]{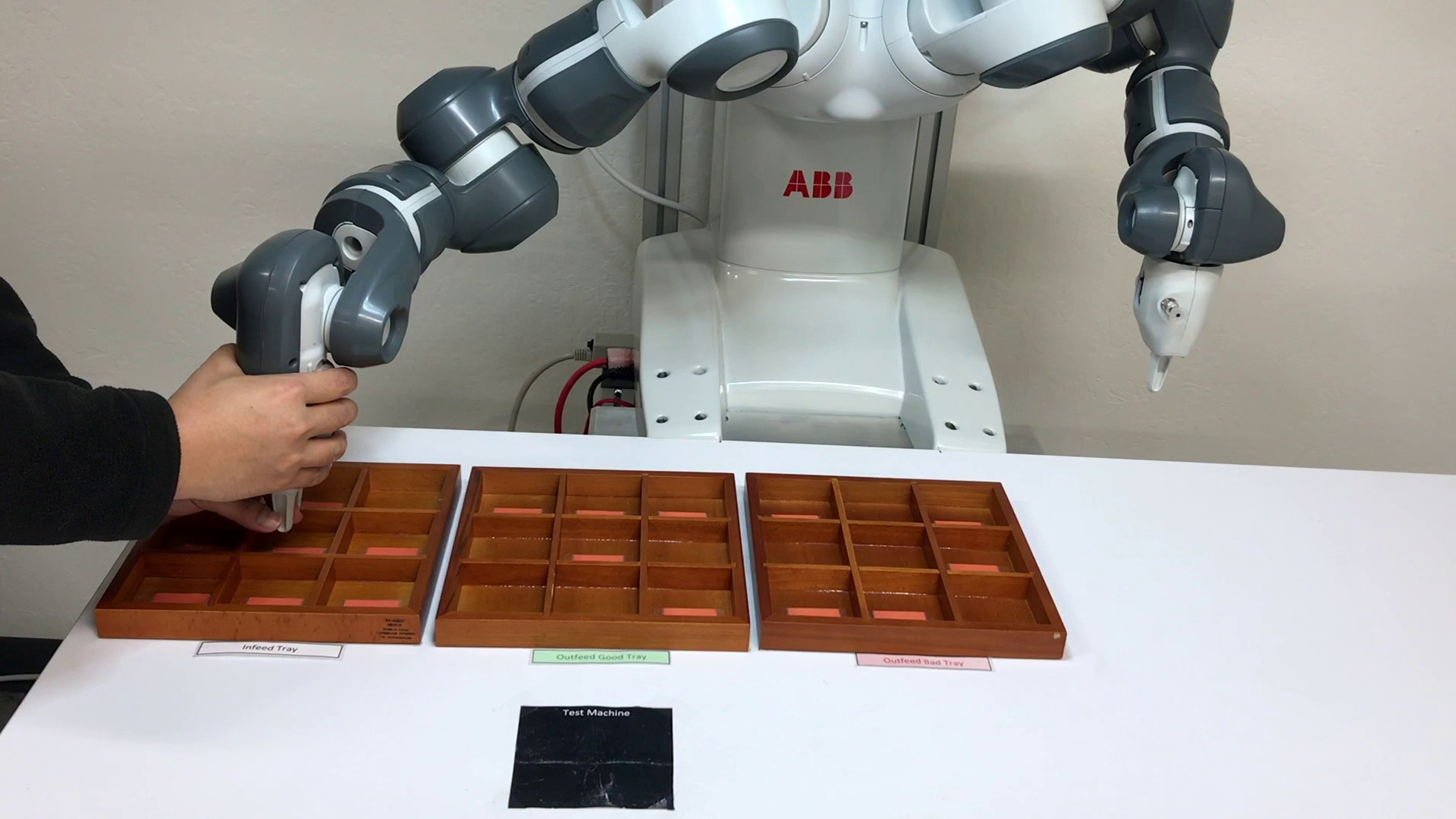}
         \caption{Tutor leads the robot to the infeed tray and closes the gripper to grasp the object.}
         \label{fig:training02}
     \end{subfigure}
     \hfill
     \begin{subfigure}[t]{0.3\textwidth}
         \centering
         \includegraphics[width=\textwidth]{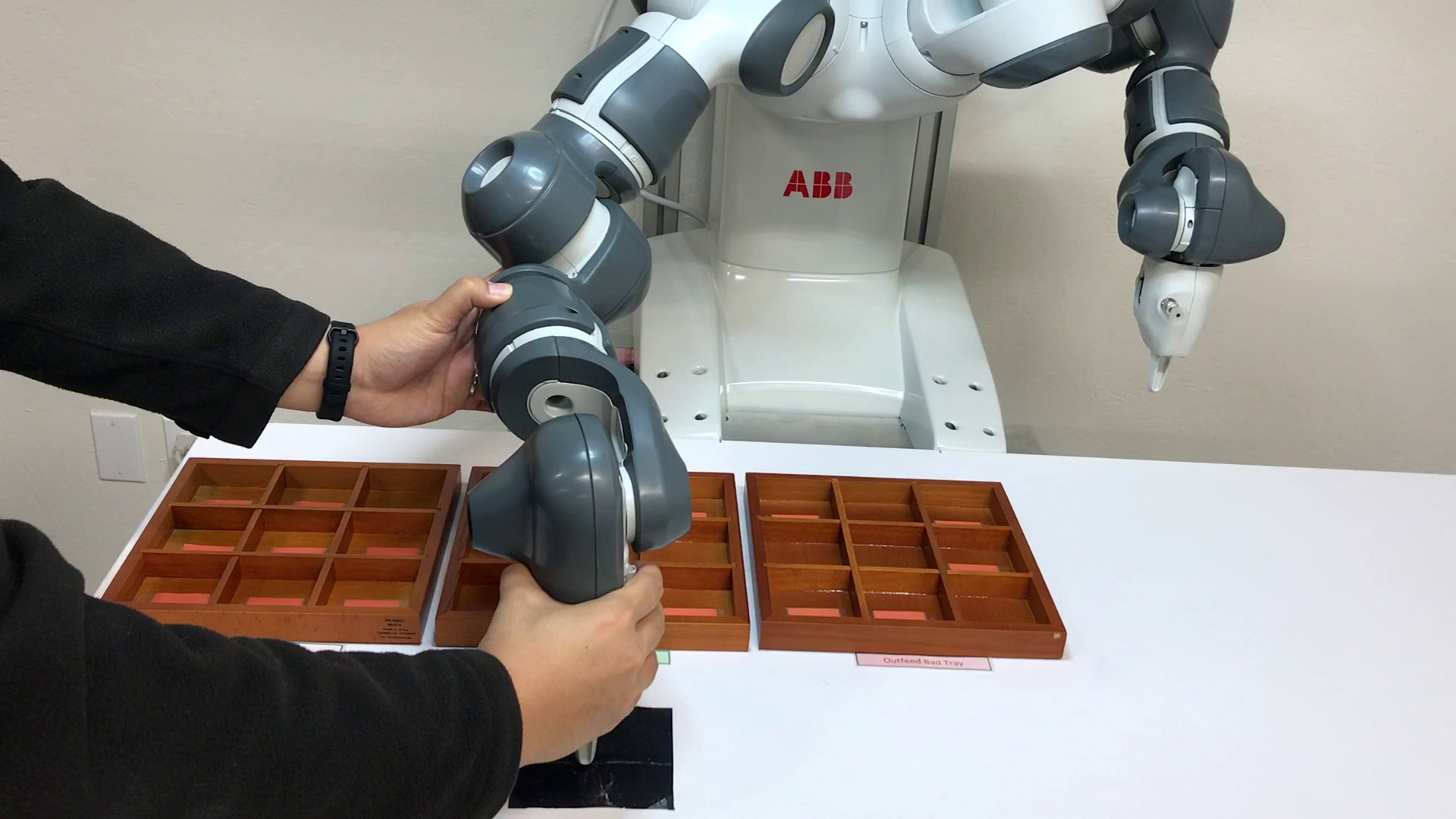}
         \caption{The robot with the closed gripper is moved to the test machine and the gripper is opened to release the object.}
         \label{fig:training03}
     \end{subfigure}
     \hfill
     
     \begin{subfigure}[t]{0.3\textwidth}
         \centering
         \includegraphics[width=\textwidth]{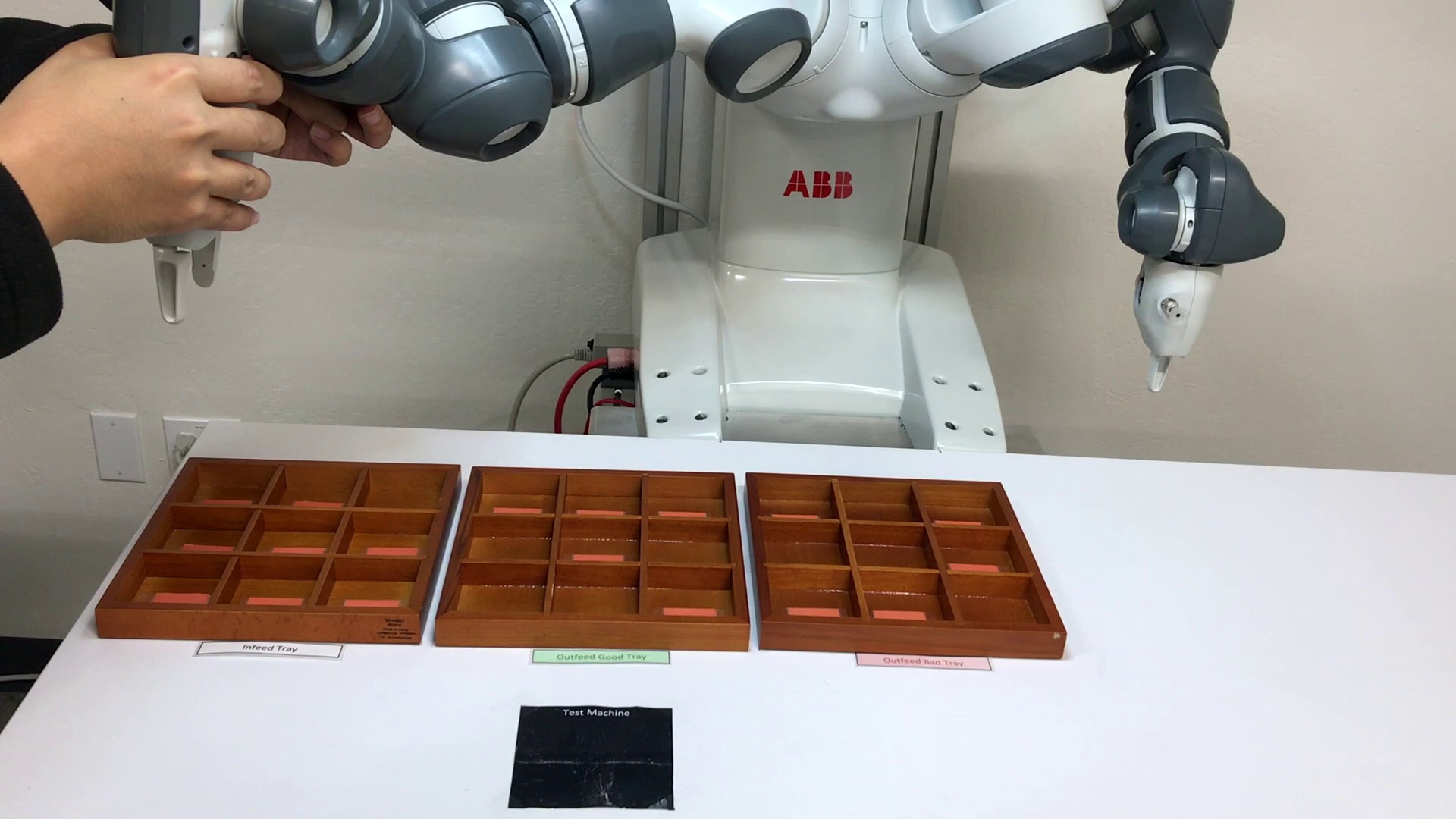}
         \caption{The robot is moved to the home position waiting for test result.}
         \label{fig:training04}
     \end{subfigure}
     \hfill
     \begin{subfigure}[t]{0.3\textwidth}
         \centering
         \includegraphics[width=\textwidth]{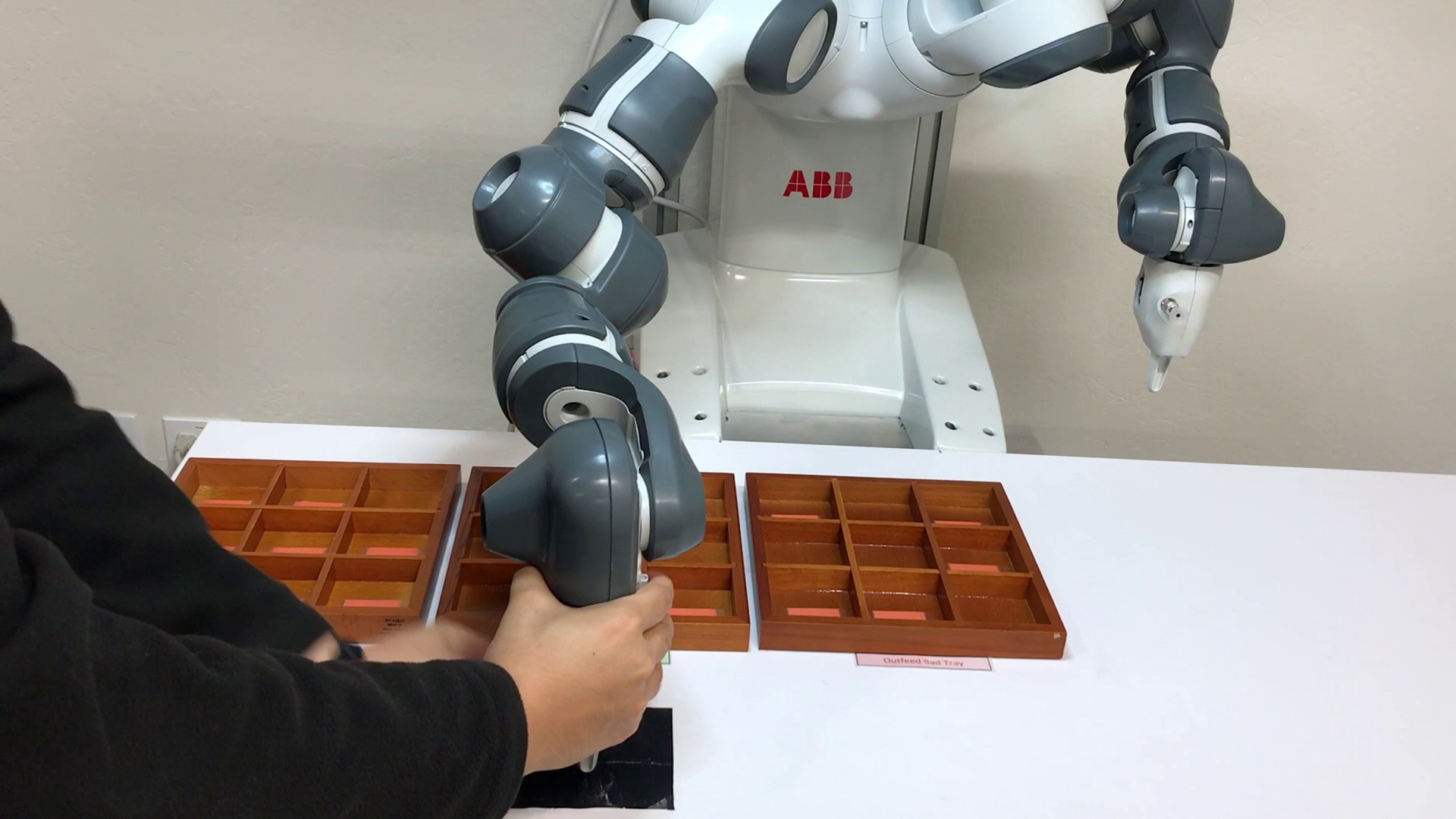}
         \caption{Post test result, the robot is moved to the test machine, gripper is closed to grasp the object.}
         \label{fig:training05}
     \end{subfigure}
     \hfill
     \begin{subfigure}[t]{0.3\textwidth}
         \centering
         \includegraphics[width=\textwidth]{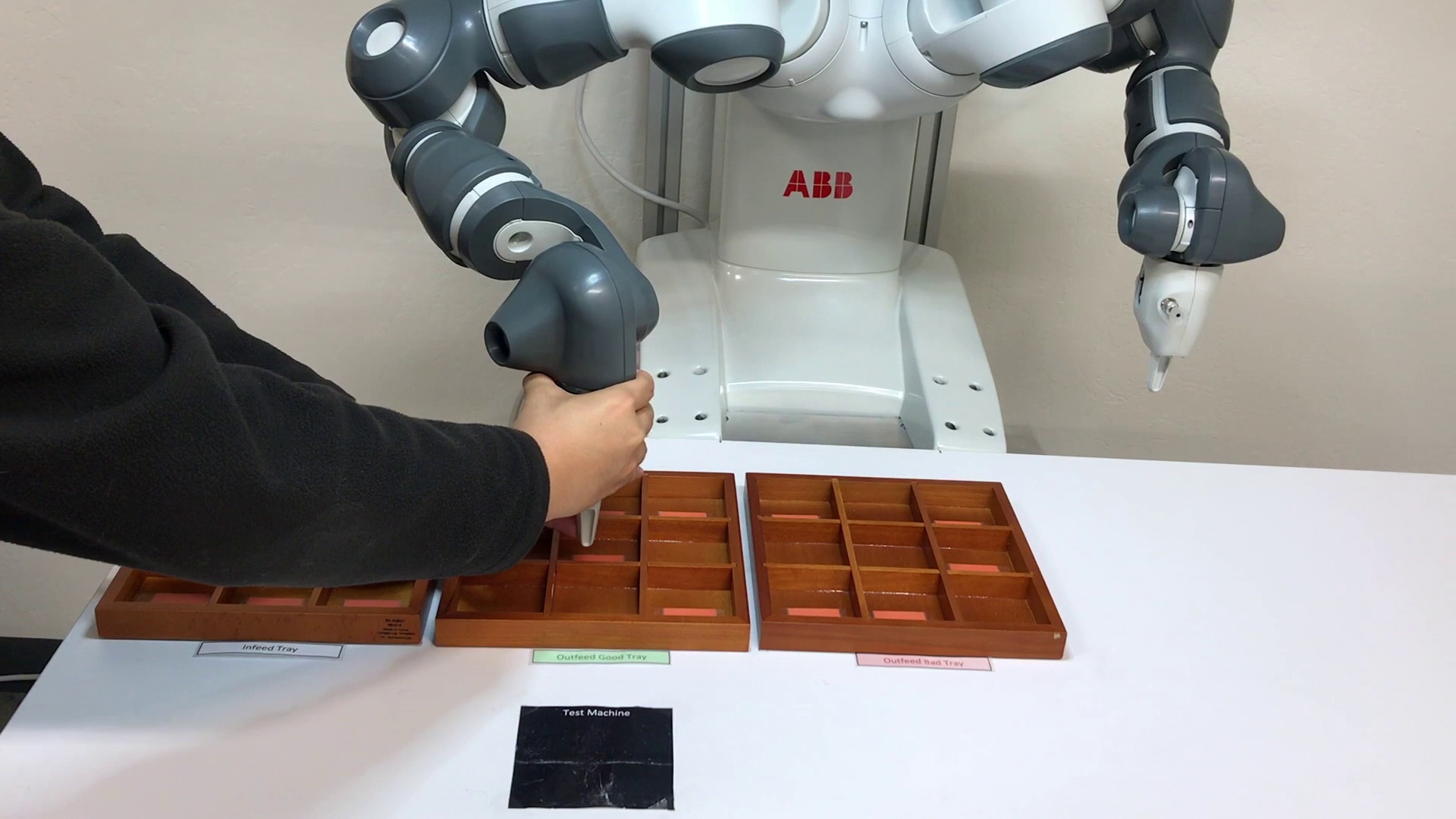}
         \caption{The robot is moved to the tray corresponding to the test result, the gripper is opened and the object is released.}
         \label{fig:training06}
     \end{subfigure}
     \hfill
     
     \begin{subfigure}[t]{0.3\textwidth}
         \centering
         \includegraphics[width=\textwidth]{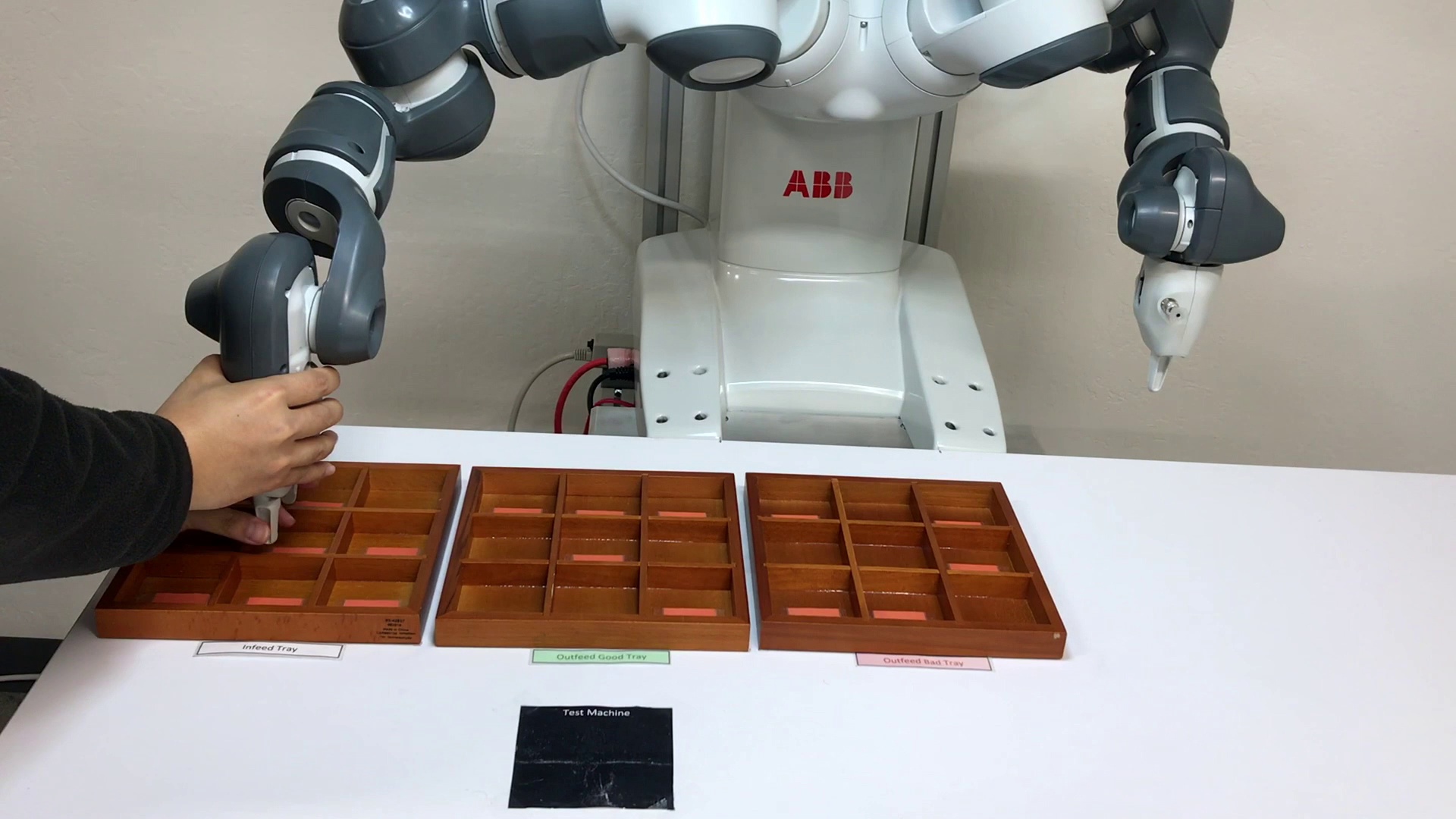}
         \caption{The robot is moved to the infeed tray to pick the next object.}
         \label{fig:training07}
     \end{subfigure}
     \hfill
     \begin{subfigure}[t]{0.3\textwidth}
         \centering
         \includegraphics[width=\textwidth]{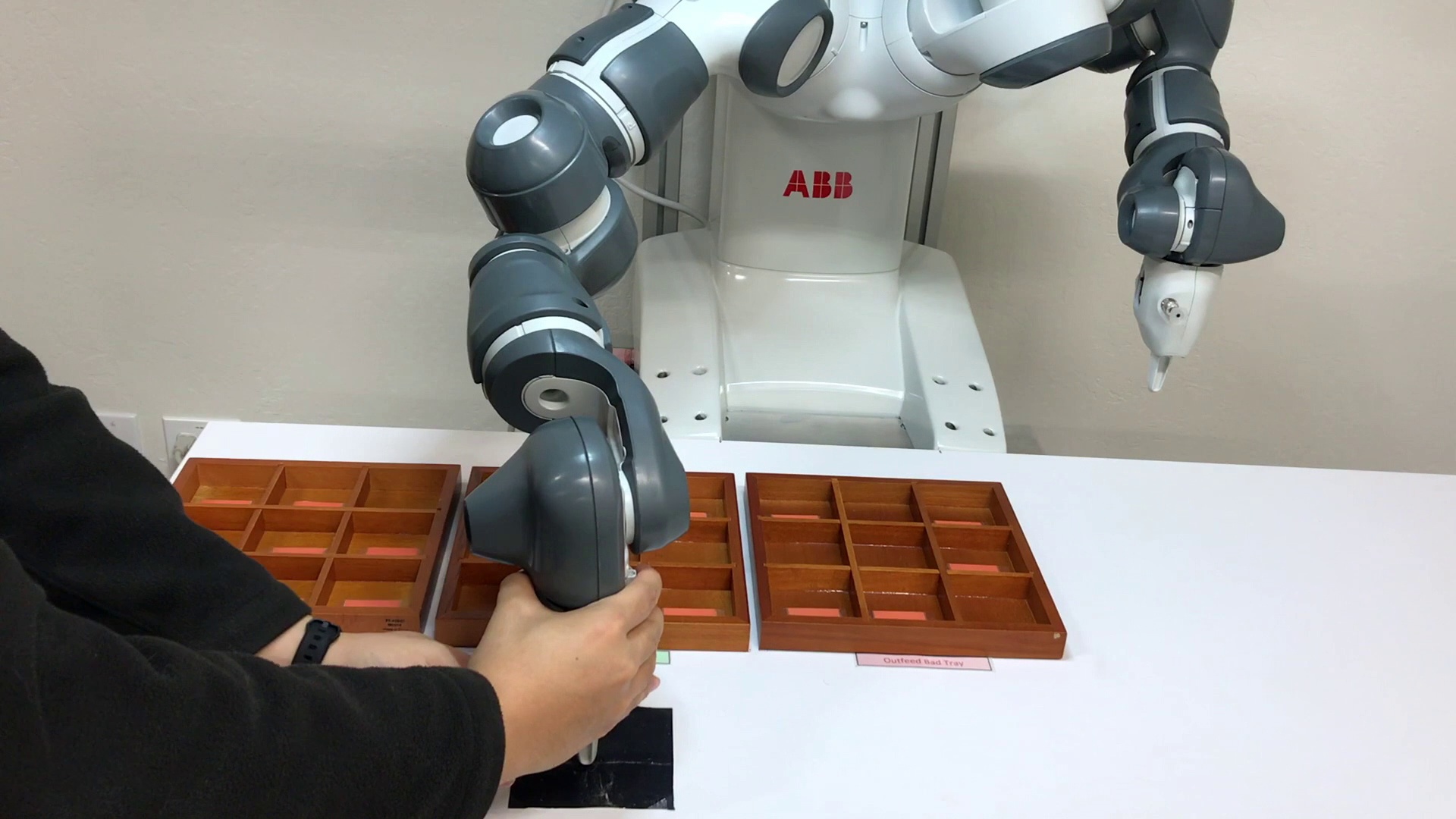}
         \caption{The object is moved to the test machine and the gripper is opened to release object.}
         \label{fig:training08}
     \end{subfigure}
     \hfill
     \begin{subfigure}[t]{0.3\textwidth}
         \centering
         \includegraphics[width=\textwidth]{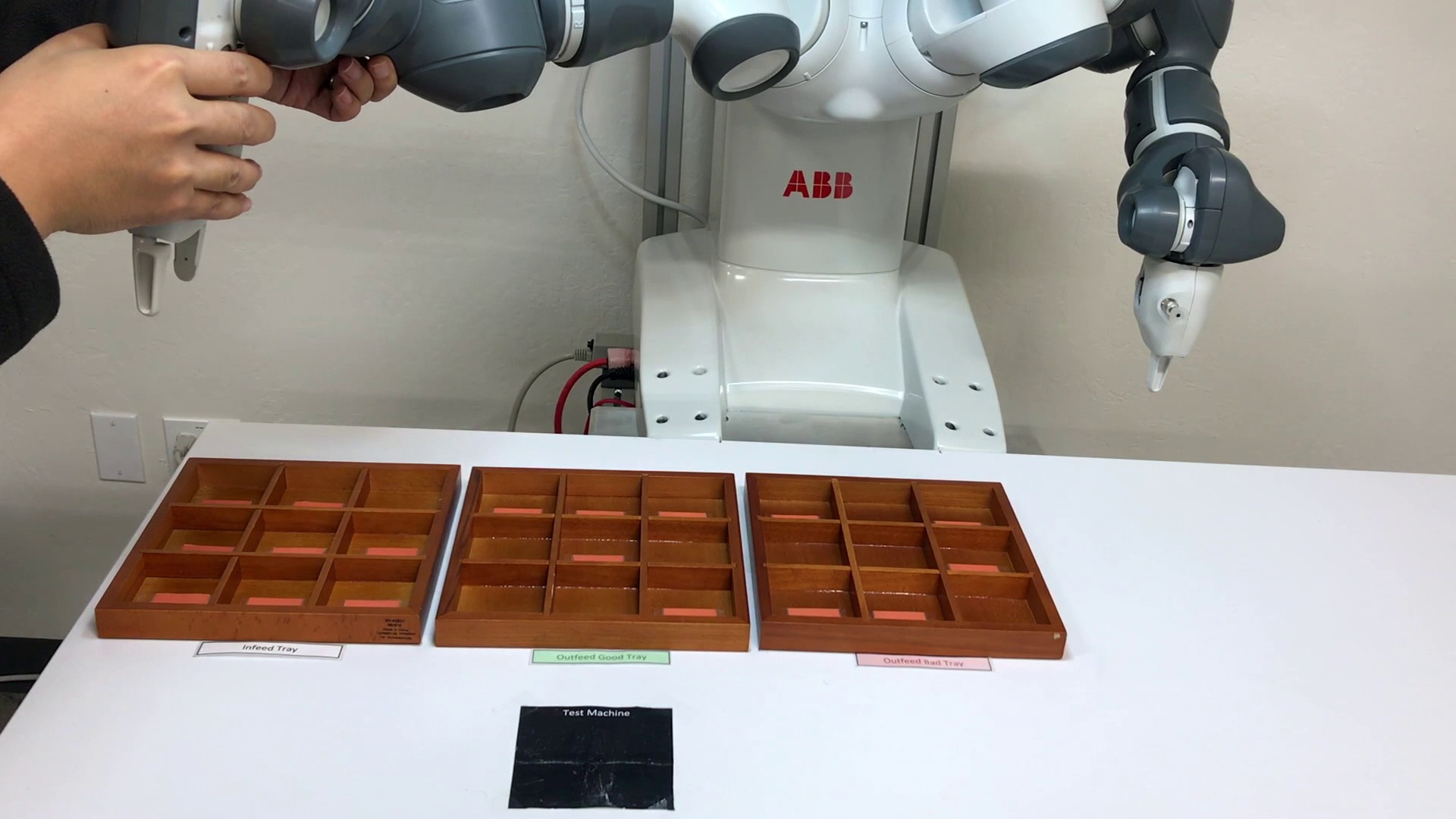}
         \caption{The robot is moved to the home position waiting for test results.}
         \label{fig:training09}
     \end{subfigure}
     \hfill
     
     \begin{subfigure}[t]{0.3\textwidth}
         \centering
         \includegraphics[width=\textwidth]{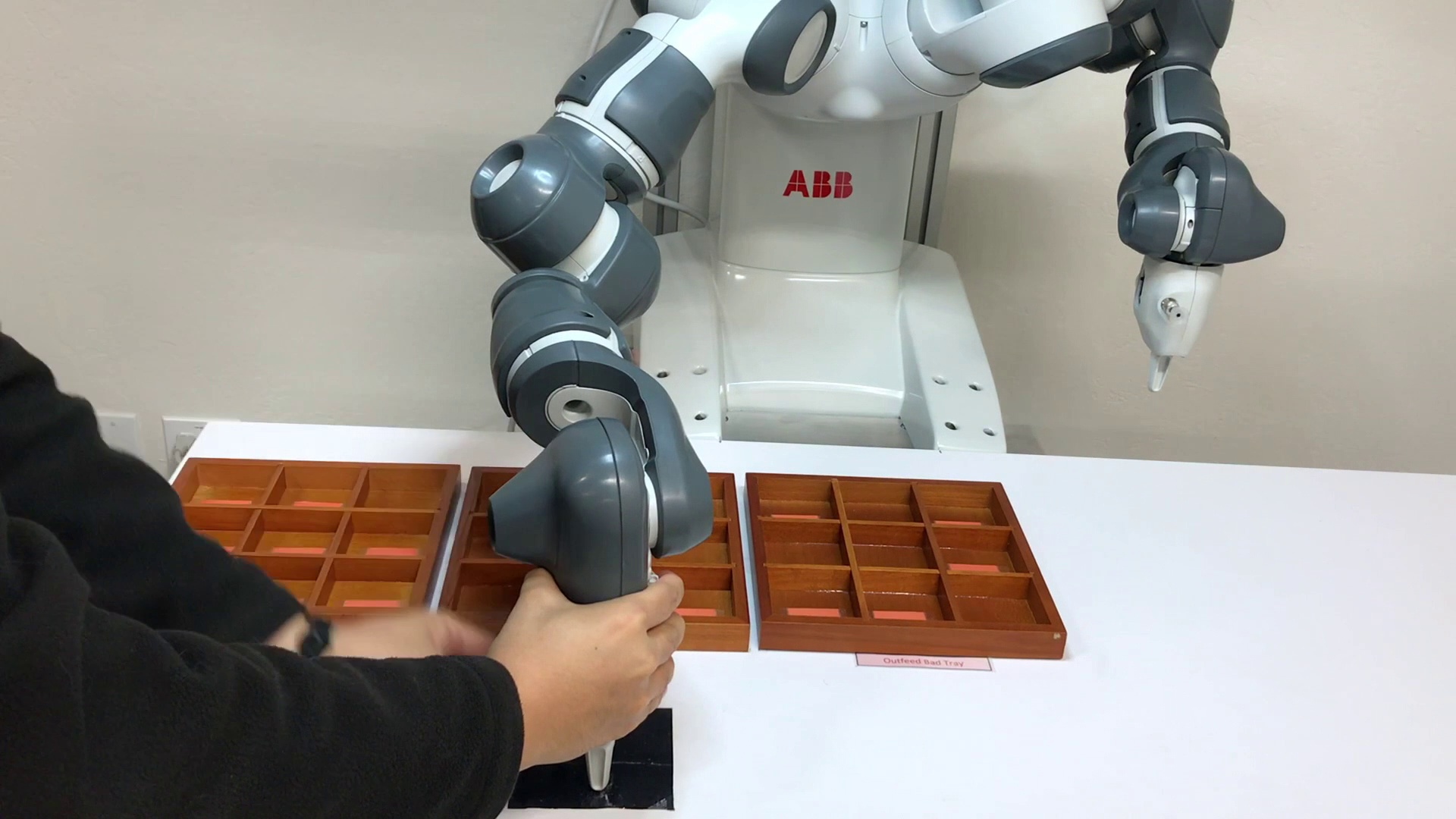}
         \caption{Once the result is knows, the object is picked from the test machine.}
         \label{fig:training10}
     \end{subfigure}
     \hfill
     \begin{subfigure}[t]{0.3\textwidth}
         \centering
         \includegraphics[width=\textwidth]{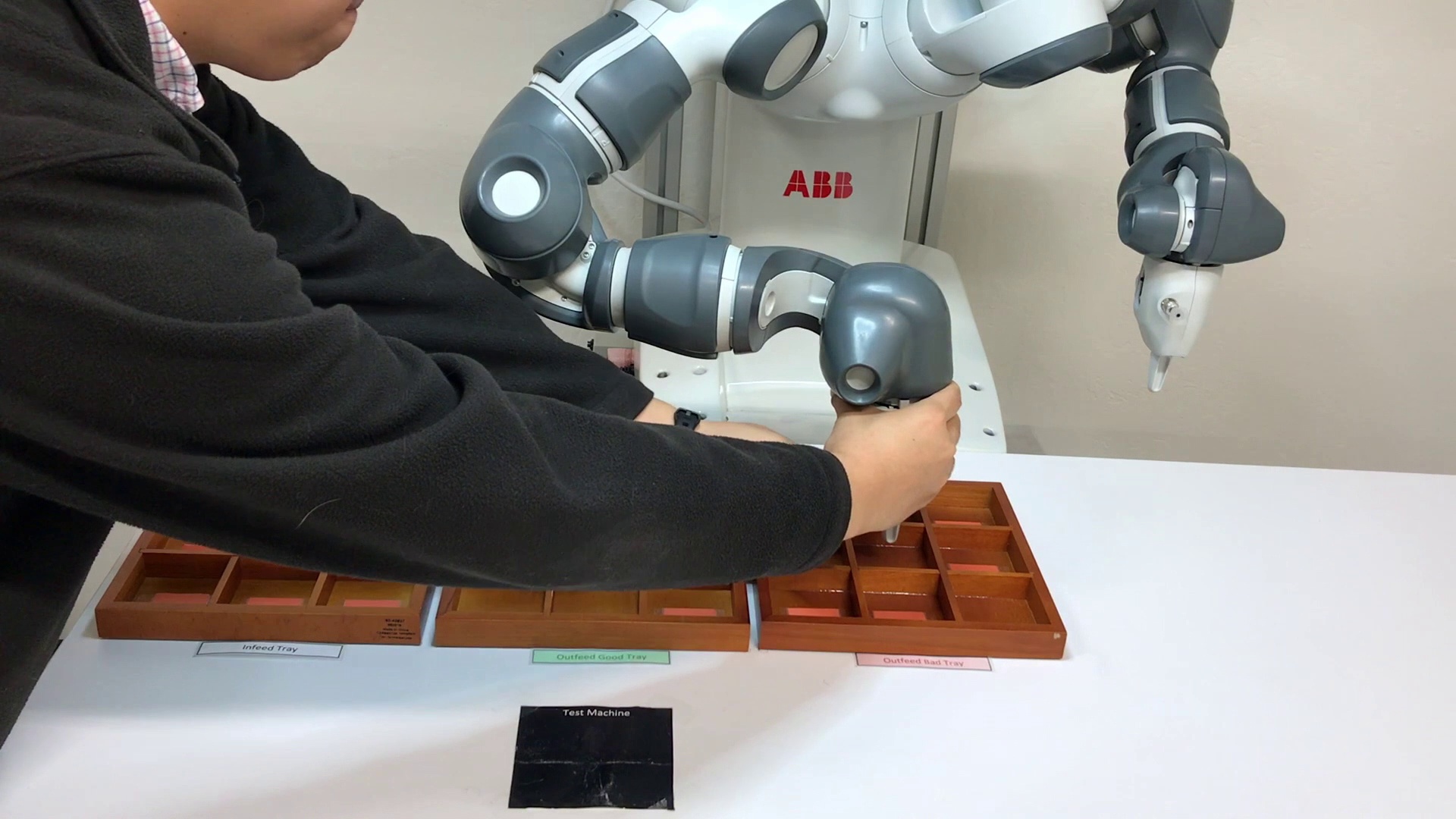}
         \caption{The object is placed in the tray corresponding to the test result.}
         \label{fig:training11}
     \end{subfigure}
     \hfill
     \begin{subfigure}[t]{0.3\textwidth}
         \centering
         \includegraphics[width=\textwidth]{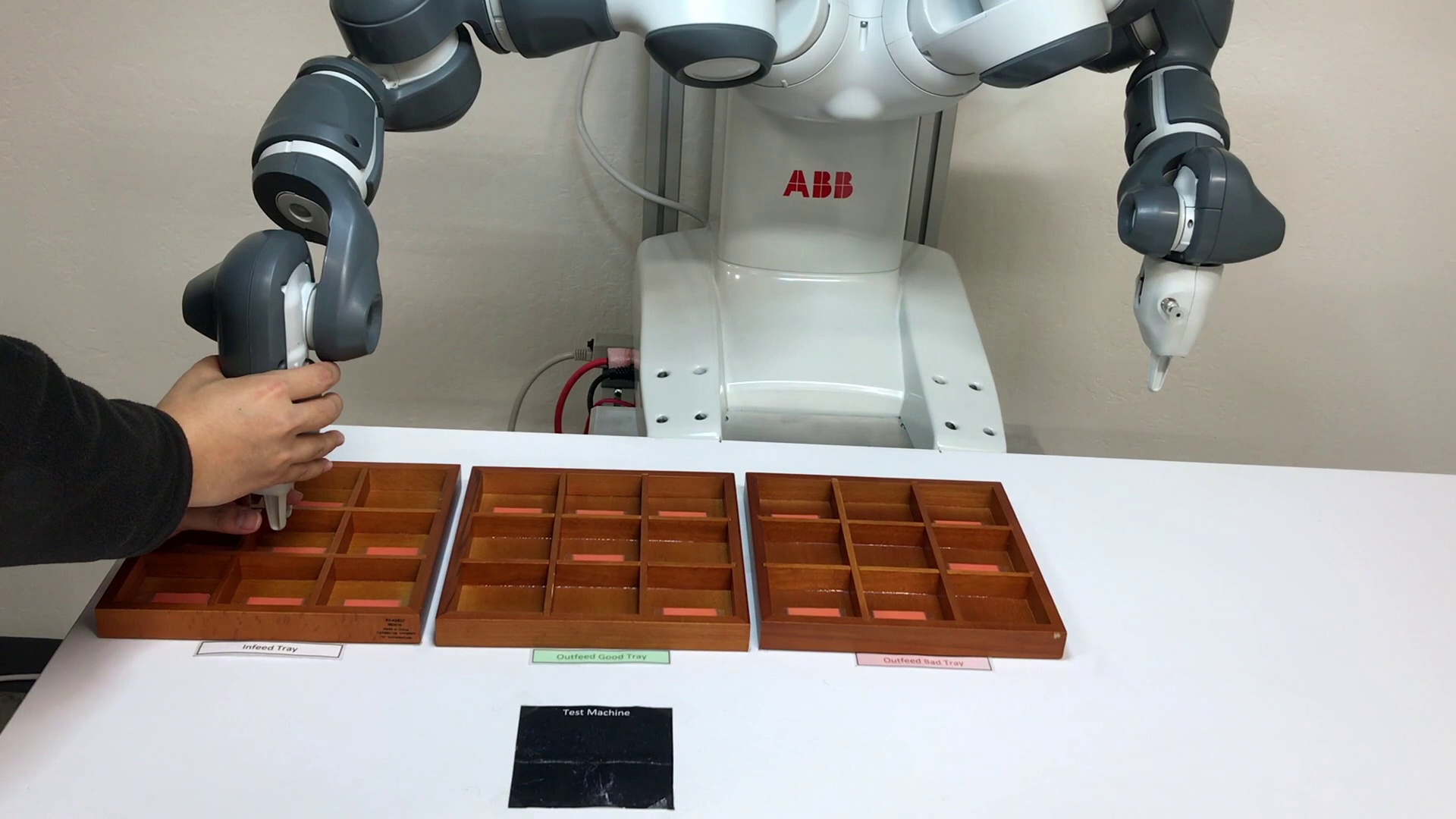}
         \caption{The robot is moved to the infeed tray to pick the next object.}
         \label{fig:training12}
     \end{subfigure}
     \hfill
     
        \caption{Training process.}
        \label{fig:training}
\end{figure*}

\begin{figure*}[t]
     \centering
     \begin{subfigure}[t]{0.3\textwidth}
         \centering
         \includegraphics[width=\textwidth]{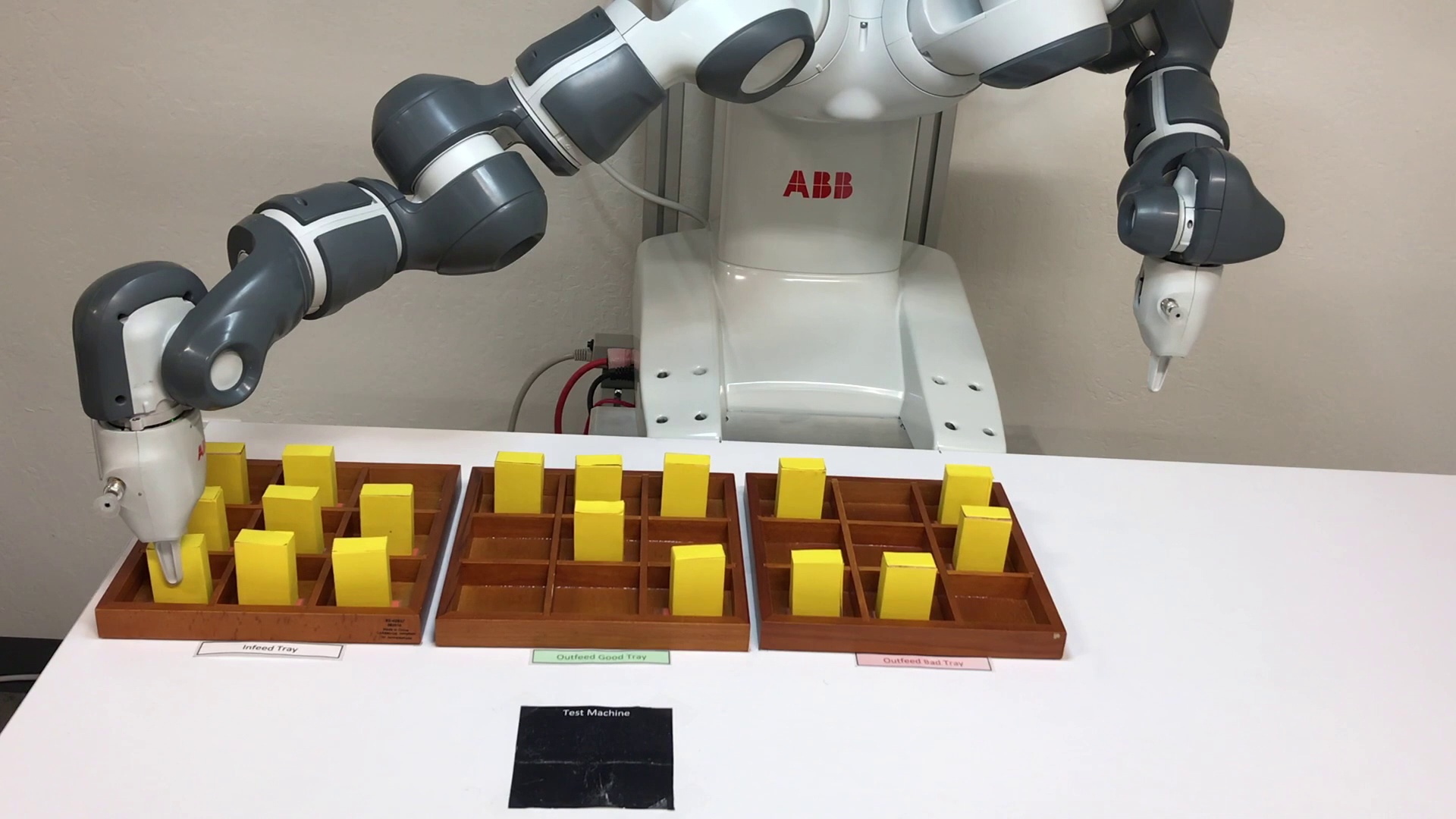}
         \caption{The robot follows the learnt process and moves to the first available object and grasps it by closing the gripper.}
         \label{fig:execution01}
     \end{subfigure}
     \hfill
     \begin{subfigure}[t]{0.3\textwidth}
         \centering
         \includegraphics[width=\textwidth]{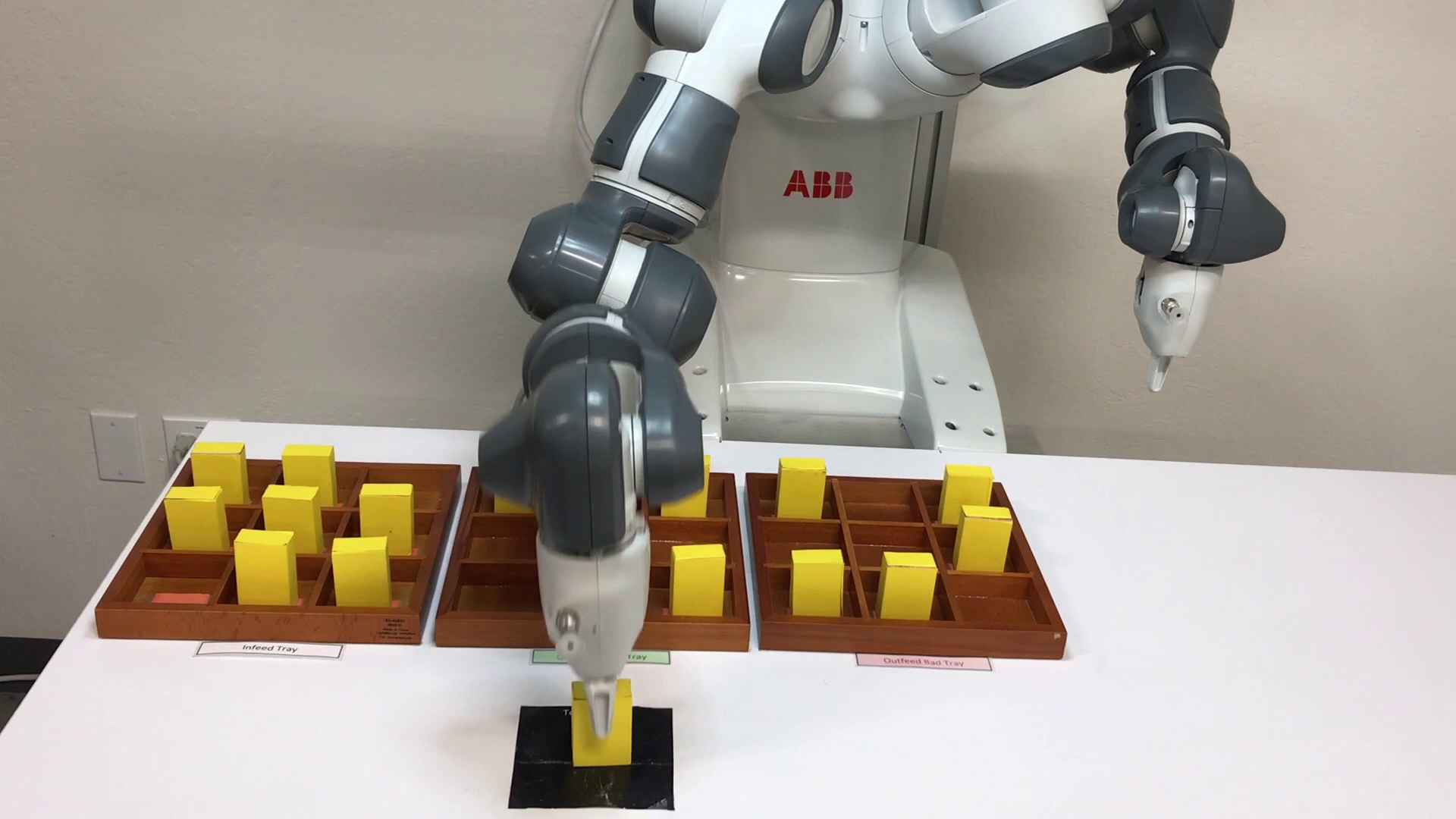}
         \caption{The object is placed in the dedicated location in the test machine and is released as the gripper is opened.}
         \label{fig:execution02}
     \end{subfigure}
     \hfill
     \begin{subfigure}[t]{0.3\textwidth}
         \centering
         \includegraphics[width=\textwidth]{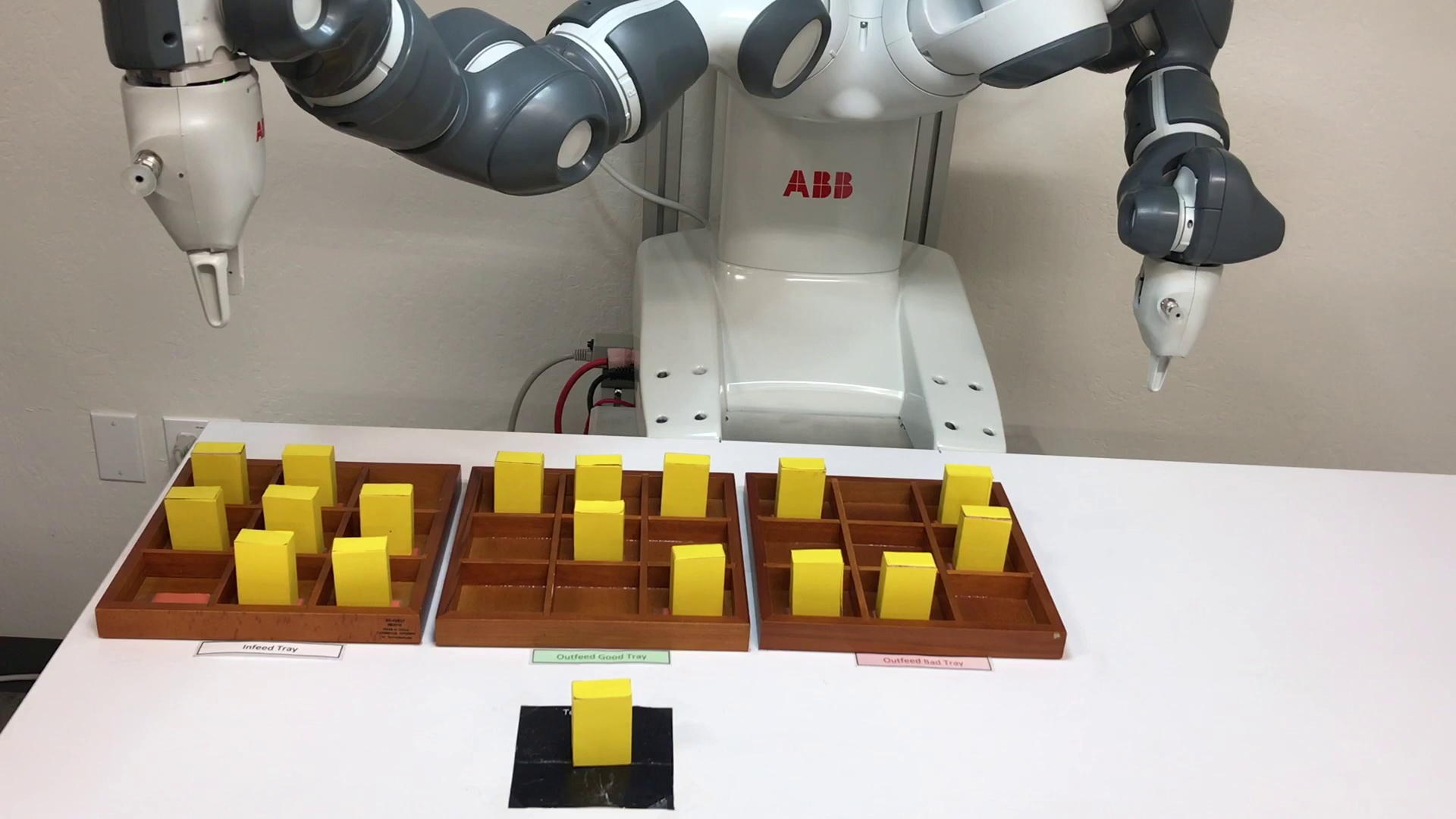}
         \caption{The robot exits the test machine and goes to the home position awaiting the test to complete.}
         \label{fig:execution03}
     \end{subfigure}
     \hfill
     
     \begin{subfigure}[t]{0.3\textwidth}
         \centering
         \includegraphics[width=\textwidth]{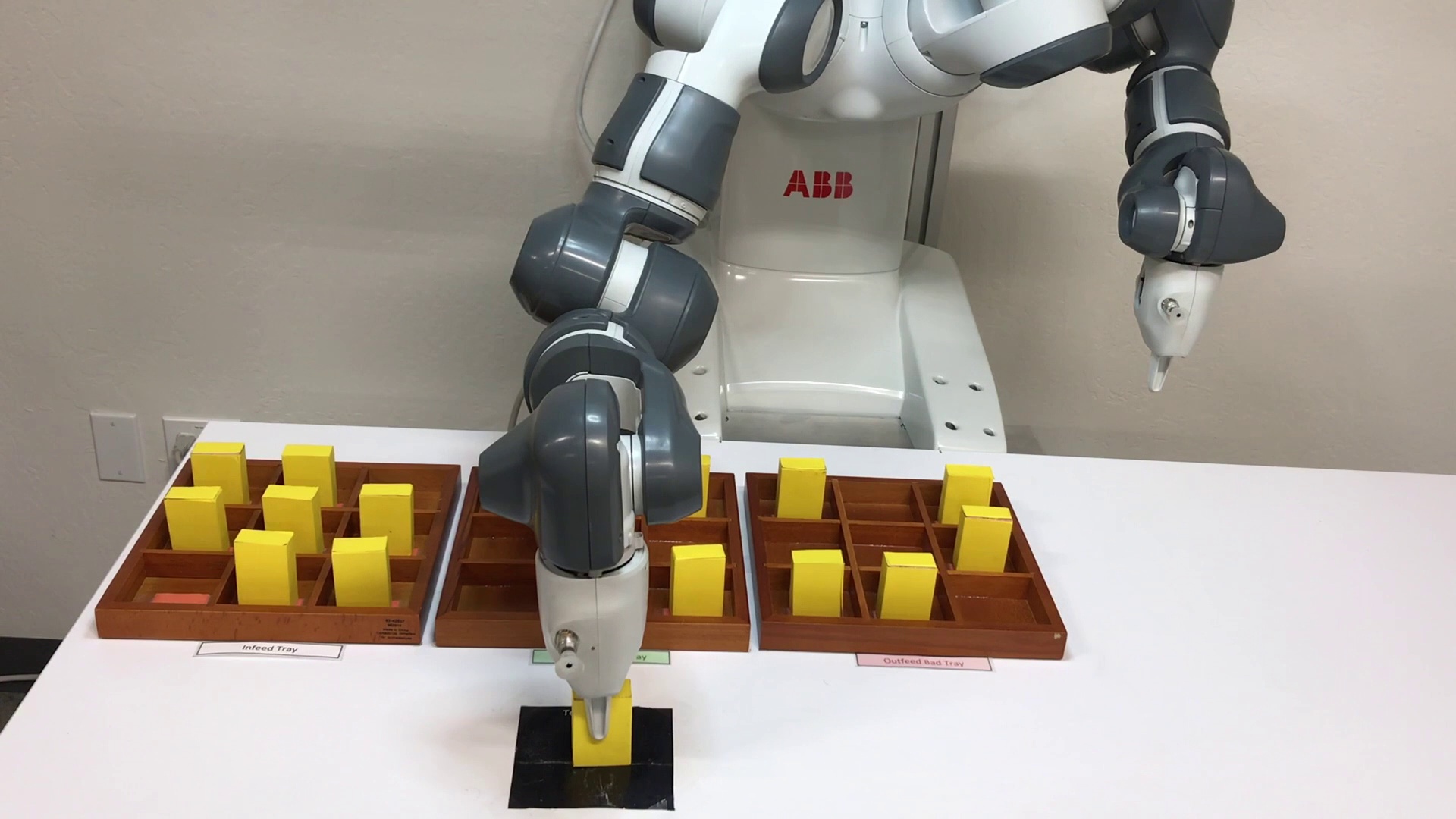}
         \caption{After that the test is complete and result is known, the robot moves to the test machine and picks the object.}
         \label{fig:execution04}
     \end{subfigure}
     \hfill
     \begin{subfigure}[t]{0.3\textwidth}
         \centering
         \includegraphics[width=\textwidth]{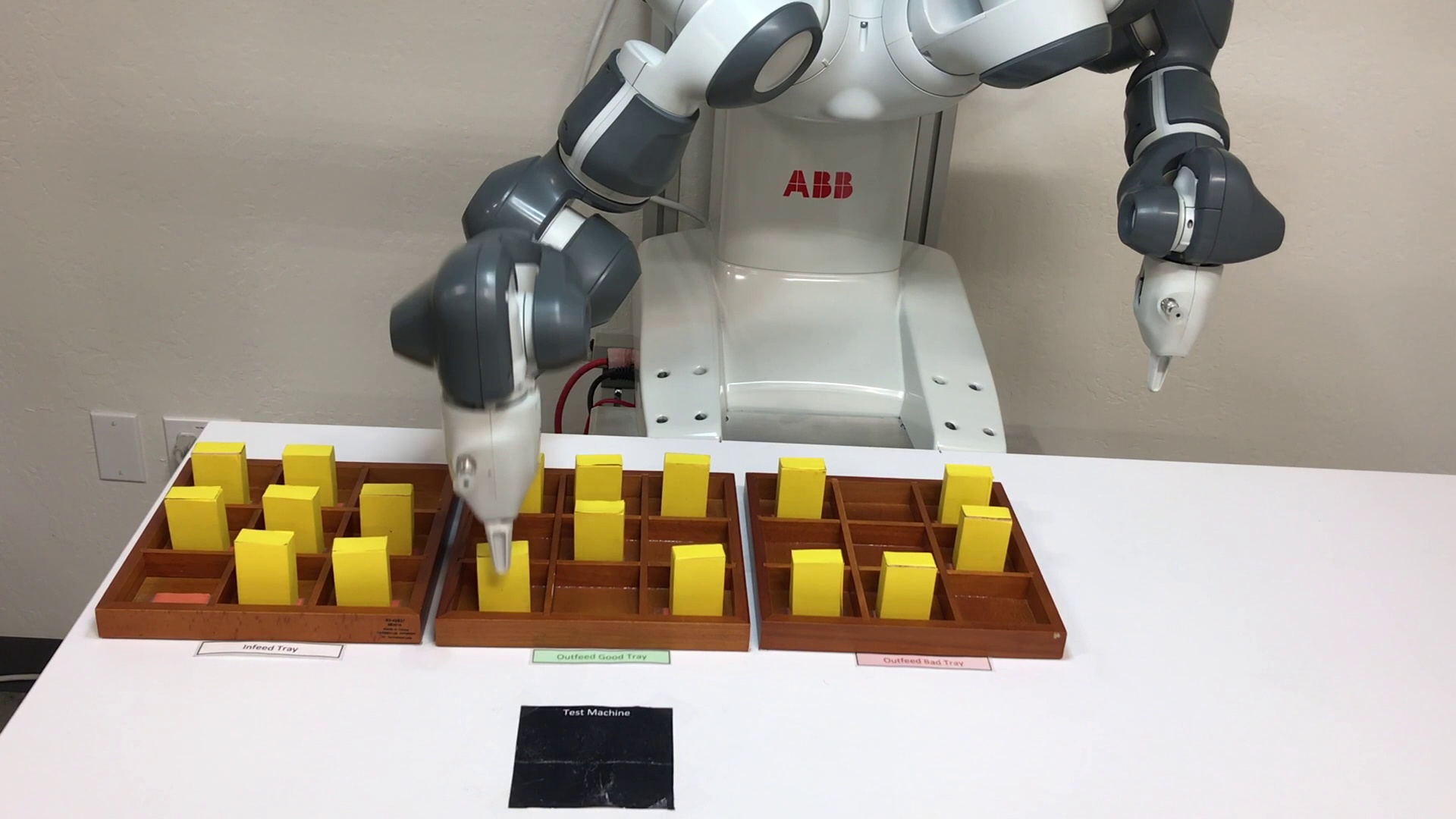}
         \caption{The target tray is selected depending on the result of the test and the object is placed in the first available position.}
         \label{fig:execution05}
     \end{subfigure}
     \hfill
     \begin{subfigure}[t]{0.3\textwidth}
         \centering
         \includegraphics[width=\textwidth]{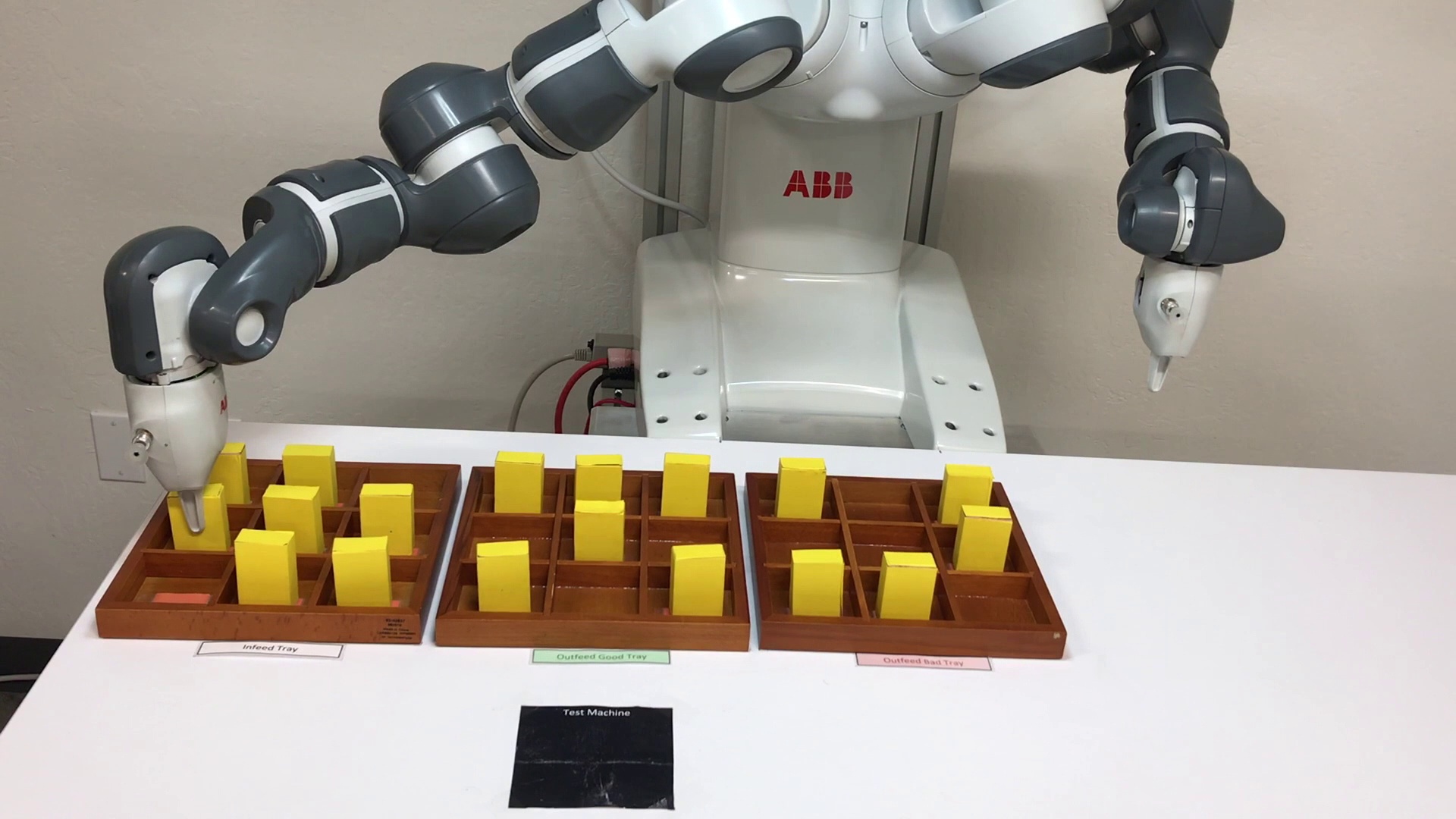}
         \caption{The robot returns to the infeed tray and picks the next available object.}
         \label{fig:execution06}
     \end{subfigure}
     \hfill
     
     \begin{subfigure}[t]{0.3\textwidth}
         \centering
         \includegraphics[width=\textwidth]{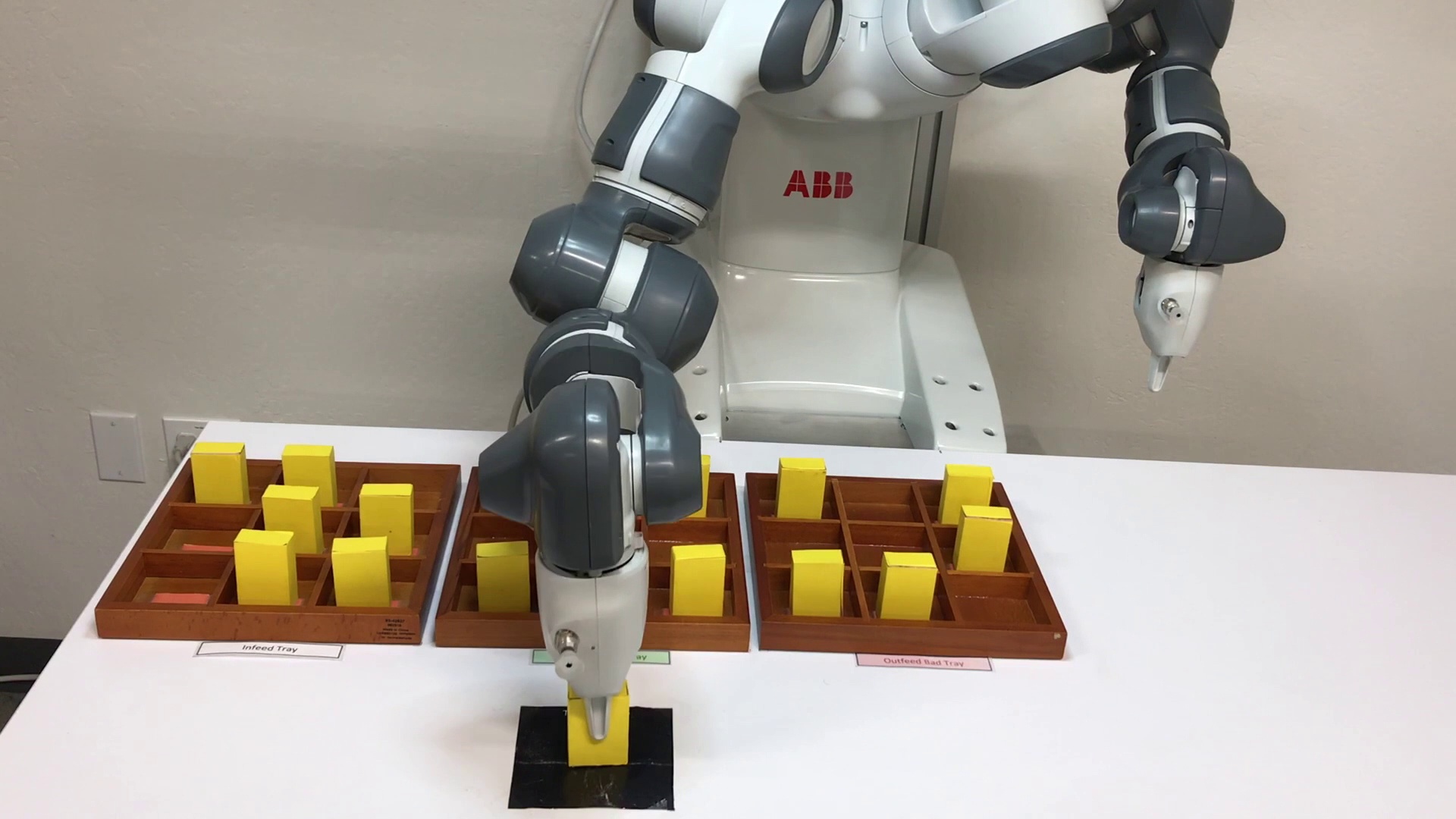}
         \caption{The robot is moved to the test machine and the object is left to be tested.}
         \label{fig:execution07}
     \end{subfigure}
     \hfill
     \begin{subfigure}[t]{0.3\textwidth}
         \centering
         \includegraphics[width=\textwidth]{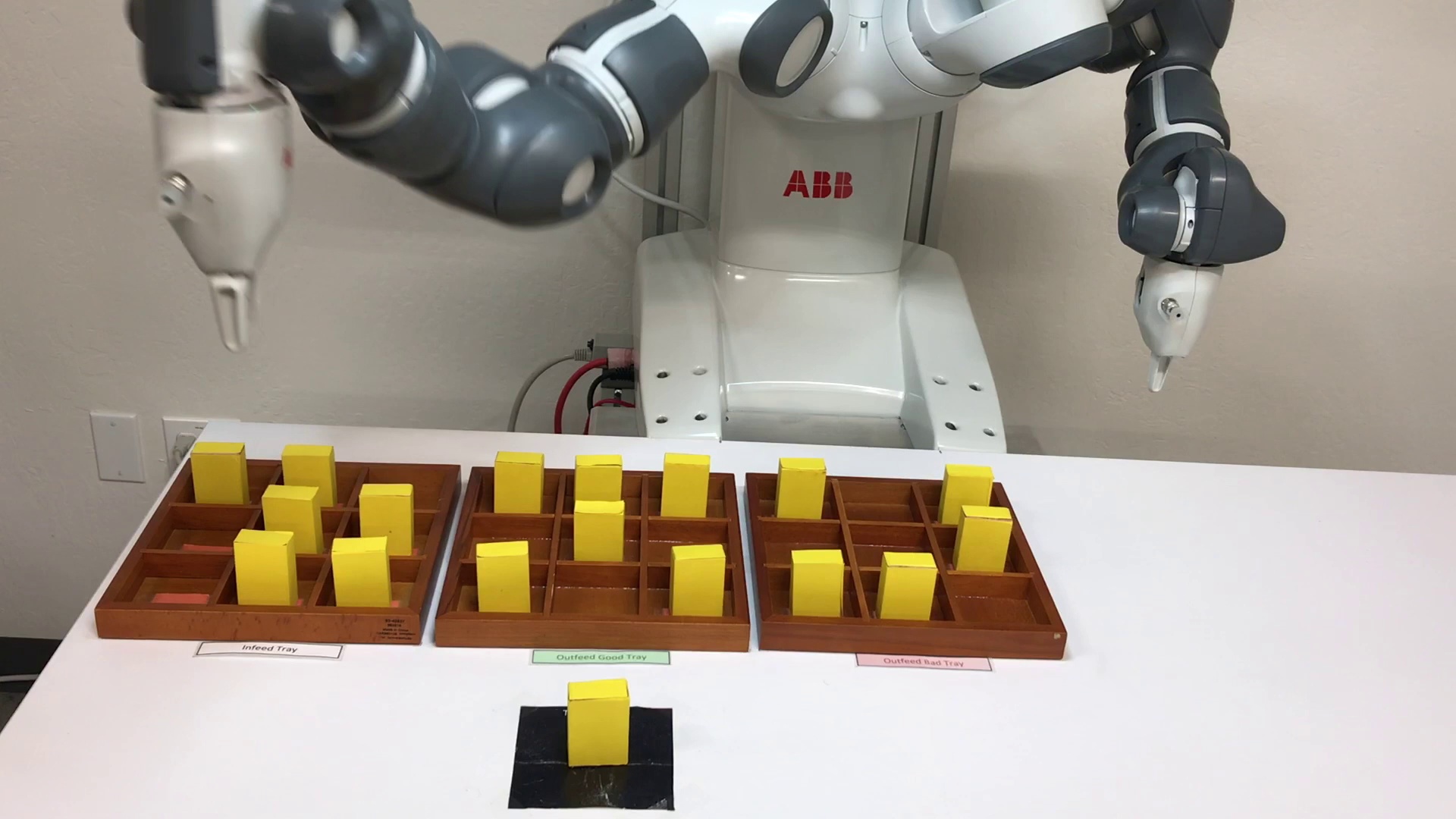}
         \caption{The robot moves to the home position waiting for the test to be completed.}
         \label{fig:execution08}
     \end{subfigure}
     \hfill
     \begin{subfigure}[t]{0.3\textwidth}
         \centering
         \includegraphics[width=\textwidth]{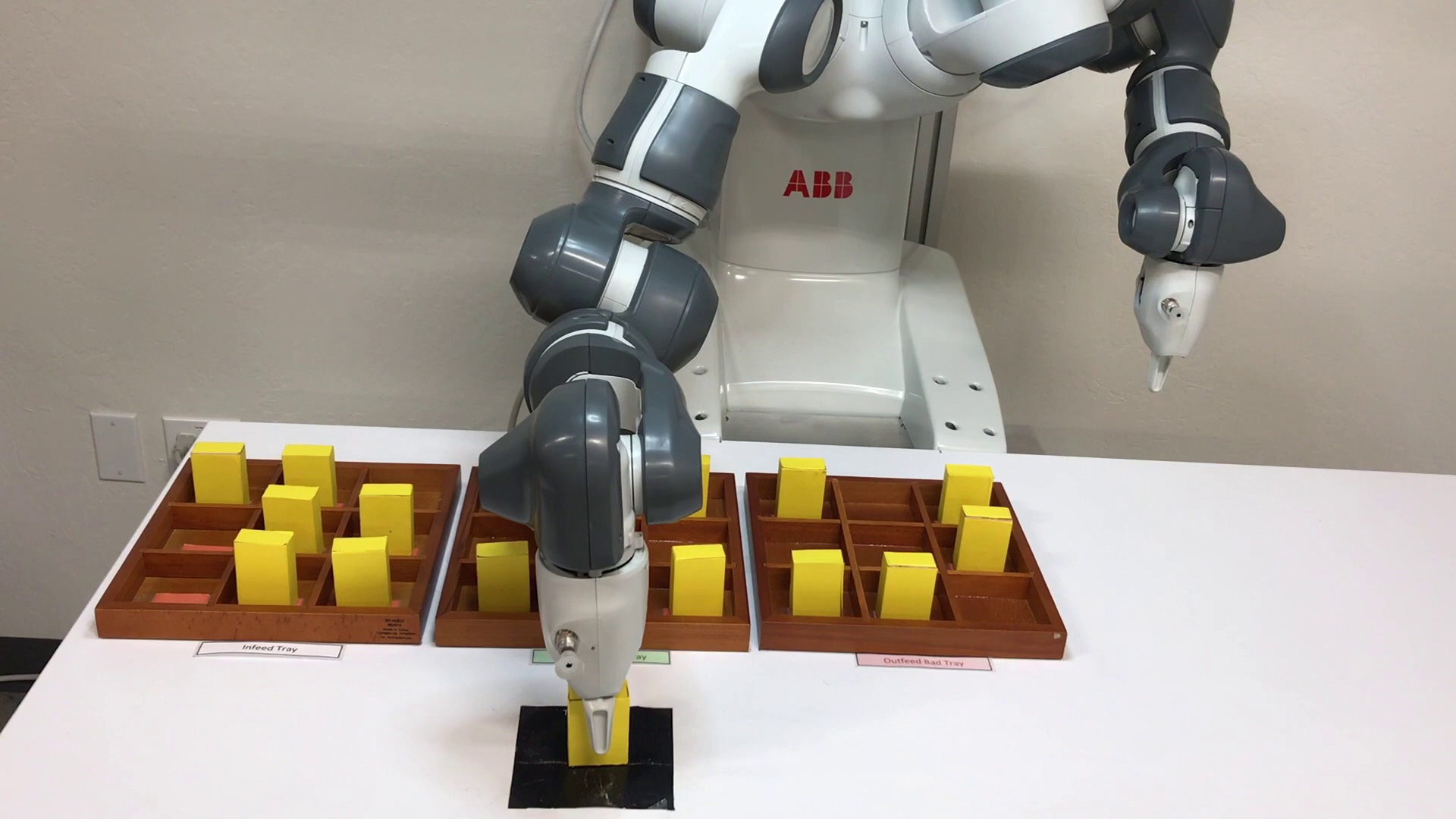}
         \caption{The robot is moved the test machine to pick the object.}
         \label{fig:execution09}
     \end{subfigure}
     \hfill
     
     \begin{subfigure}[t]{0.3\textwidth}
         \centering
         \includegraphics[width=\textwidth]{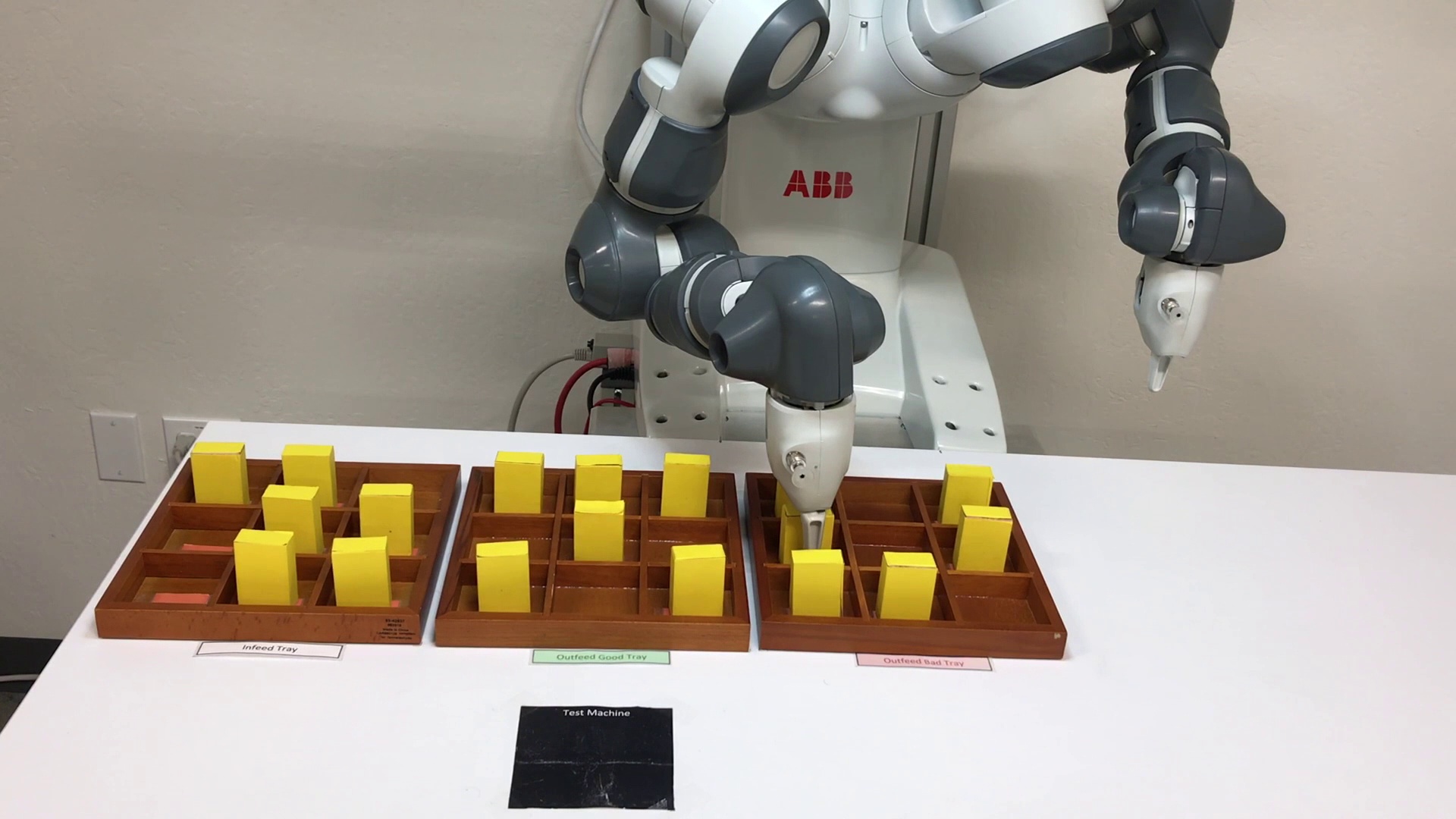}
         \caption{The object is placed in the tray corresponding to the test result.}
         \label{fig:execution10}
     \end{subfigure}
     \hfill
     \begin{subfigure}[t]{0.3\textwidth}
         \centering
         \includegraphics[width=\textwidth]{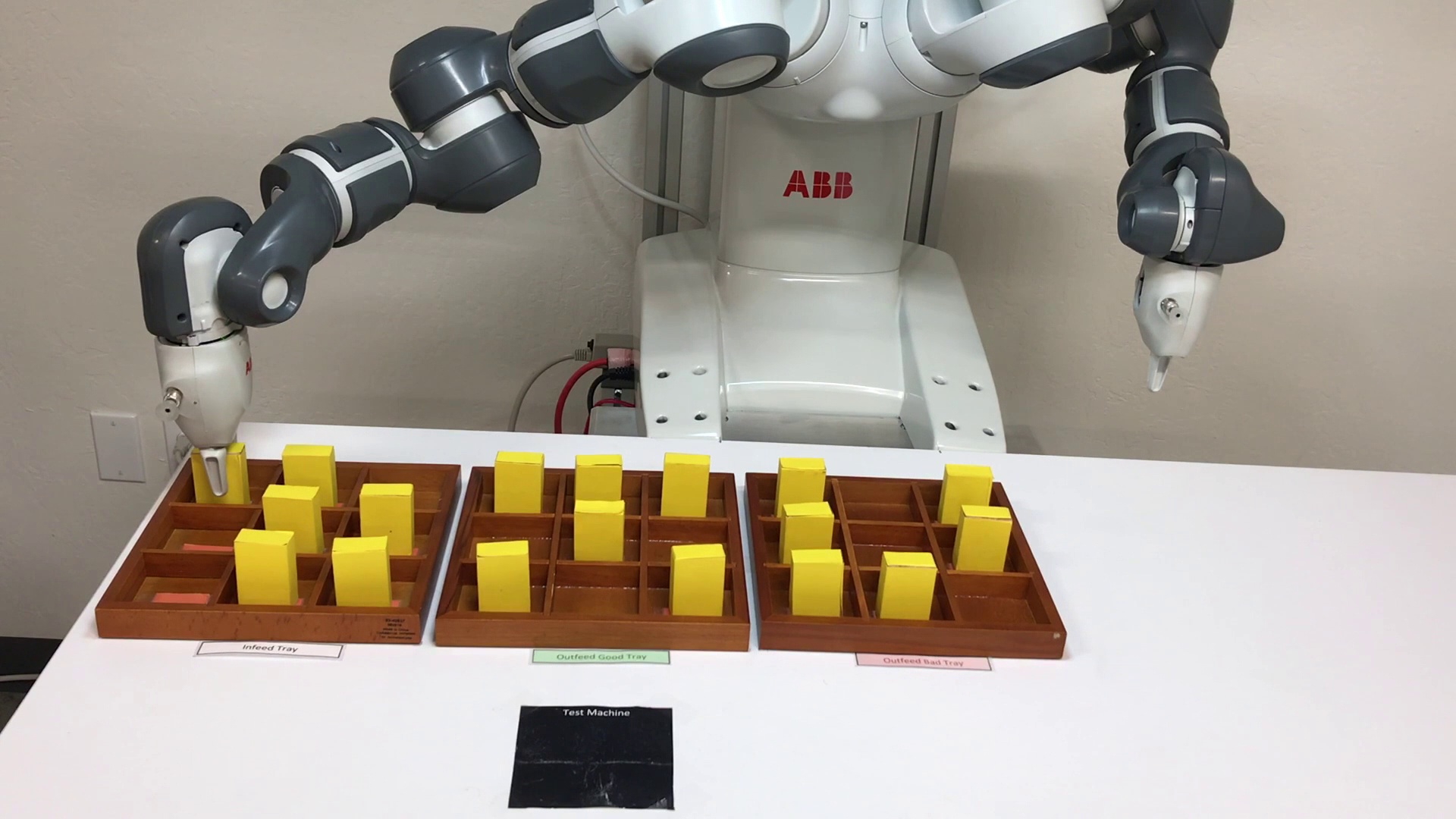}
         \caption{The robot moves to the infeed tray to pick the next object.}
         \label{fig:execution11}
     \end{subfigure}
     \hfill
     \begin{subfigure}[t]{0.3\textwidth}
         \centering
         \includegraphics[width=\textwidth]{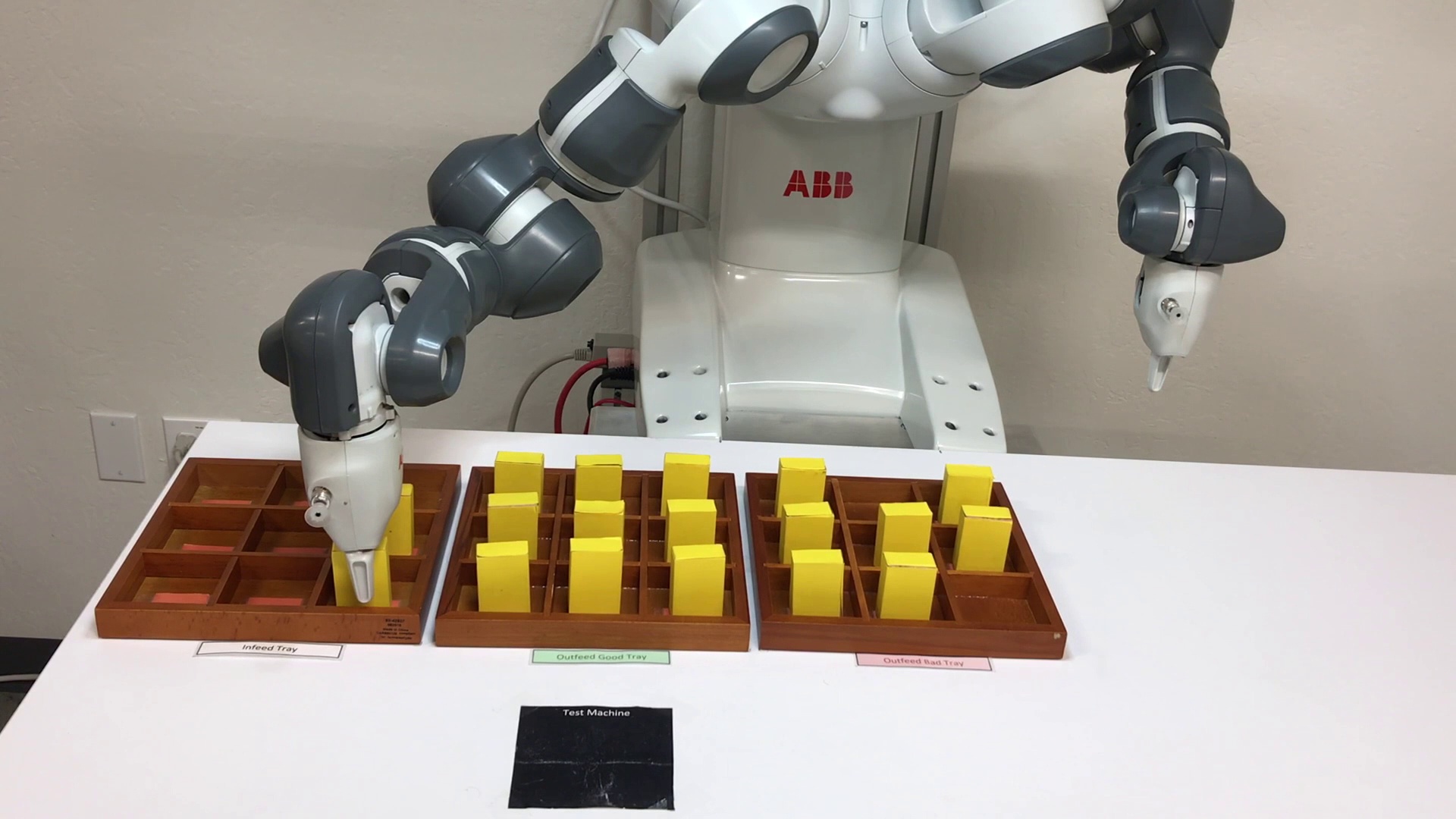}
         \caption{The process is continued until an unforeseen state occurs. Here, the middle tray is full, and therefore the robot stops and waits for human intervention.}
         \label{fig:execution12}
     \end{subfigure}
     \hfill
     
        \caption{Execution process.}
        \label{fig:execution}
\end{figure*}

Each recognized element is expected to be represented by a state machine as described above. That is all information which will be available to the system before training. The information is generic and does not depend on the application to be trained. Consequently, before training the robot system does not have any information about what the task will be and how it should be performed. However, after recognizing the elements in the system and in the environment, the robot is capable to record any changes and can learn to react to them according to the instructions from the tutor.

We assume that the process will start with placing a full infeed tray in the scene. This change of environment state will be detected and will trig execution of the process that has been learnt. In what follows, we describe the training process first and then show how the robot will repeat the process accurately. Once again, it should be noted that the robot does not learn a path, but understands the semantic of the actions, including all logic.

Figure ~\ref{fig:training} presents the steps that the tutor needs to take to teach the process to the robot system. The tray to the left is the infeed and the middle one and the right one are outfeed trays associated to the binary result of the test, say 'good' and 'defected'. As described in the captions of the sub-figures, the tutor simply needs to guide the robot to perform the task at least once for each possible branch, here a good object and a bad object. Obviously, during the training time, it would be practical to have one good and one bad at the start and let the robot continue by itself.

Actions that are taken by the tutor, or autonomously by the elements themselves, are observed and used for dividing the process into meaningful steps that should be executed when certain well-defined conditions are fulfilled. These steps are assembled as the application state machine that can be executed in continuation.

Figure ~\ref{fig:execution} shows how the robot repeats the steps that it has learnt after a complete training session. Obviously, to be able to perform the work, the robot needs to understand the semantics and generalize the learnt step. For example, at the pick, the robot picks next available object, whose coordinate will be delivered by the tray, which tracks its content with the help of the camera. Equally, at the placement, the next available empty slot is taken, without the tutor having taught that specific position.

The last sub-figure in figure ~\ref{fig:execution}, figure ~\ref{fig:execution12}, shows that the robot has observed a yet unknown state: The middle tray is full, an event that has not occurred before. In such a case, the system lacks information to proceed and will stop and request human interventions. The operator will take action, for example instruct the other hand to remove the full tray and replace it with an empty one. This will bring the system to a recognized state and the procedure can continue. By doing so, how to handle exceptions can also be learned after that each exception has happened only once. This reduces the burden of programming enormously. 



\section{Summary and conclusion}



We propose an approach that allows machines to record and follow instructions in a reliable manner. Each element of the system is modeled by a state machine that describes the behavior of the element and its interaction with other elements in the system. The element model is independent of the application, allowing its reuse across multiple applications. The training process consists of a single demonstration of the tasks, allowing the system to comprehend the steps and the underlying inherent logic.

This procedure minimizes programming efforts and automatically generates the execution logic of the tasks performed. Occurrence of untrained circumstances leads to the system prompting for unexpected state and requesting human intervention for complementary training. Episodic memory based design leads to continuous learning for robots by exploiting effective memory structures that utilizes past experiences.

\bibliographystyle{IEEEtran}
\typeout{}
\bibliography{IEEEexample}

\begin{thebibliography}{10}
\providecommand{\url}[1]{#1}
\csname url@samestyle\endcsname
\providecommand{\newblock}{\relax}
\providecommand{\bibinfo}[2]{#2}
\providecommand{\BIBentrySTDinterwordspacing}{\spaceskip=0pt\relax}
\providecommand{\BIBentryALTinterwordstretchfactor}{4}
\providecommand{\BIBentryALTinterwordspacing}{\spaceskip=\fontdimen2\font plus
\BIBentryALTinterwordstretchfactor\fontdimen3\font minus
  \fontdimen4\font\relax}
\providecommand{\BIBforeignlanguage}[2]{{%
\expandafter\ifx\csname l@#1\endcsname\relax
\typeout{** WARNING: IEEEtran.bst: No hyphenation pattern has been}%
\typeout{** loaded for the language `#1'. Using the pattern for}%
\typeout{** the default language instead.}%
\else
\language=\csname l@#1\endcsname
\fi
#2}}
\providecommand{\BIBdecl}{\relax}
\BIBdecl

\bibitem{ref01}
S.~Mitsi, K.-D. Bouzakis, G.~Mansour, D.~Sagris, and G.~Maliaris, ``Off-line
  programming of an industrial robot for manufacturing,'' \emph{The
  International Journal of Advanced Manufacturing Technology}, 2005.

\bibitem{ref02}
Z.~Pan, J.~Polden, N.~Larkin, S.~V. Duin, and J.~Norrish, ``Recent progress on
  programming methods for industrial robots,'' \emph{Robotics and
  Computer-Integrated Manufacturing}, 2012.

\bibitem{ref03}
M.~A. Potter, K.~A.~D. Jong, and J.~J. Grefenstette, ``A coevolutionary
  approach to learning sequential decision rules,'' \emph{ICGA}, 1995.

\bibitem{ref04}
S.~Niekum, S.~Osentoski, G.~Konidaris, and A.~G. Barto, ``Learning and
  generalization of complex tasks from unstructured demonstrations,''
  \emph{IROS}, 2012.

\bibitem{ref05}
A.~M. Howard, C.~H. Park, and S.~Remy, ``Using haptic and auditory interaction
  tools to engage students with visual impairments in robot programming
  activities,'' \emph{IEEE transactions on Learning Technologies}, 2012.

\bibitem{ref06}
U.~Thomas, G.~Hirzinger, B.~Rumpe, C.~Schulze, and A.~Wortmann, ``A new skill
  based robot programming language using uml/p statecharts,'' \emph{ICRA},
  2013.

\bibitem{ref07}
X.~Long and T.~Padır, ``Template-based human supervised robot task
  programming,'' \emph{IROS}, 2016.

\bibitem{ref08}
P.~T. Cox and T.~J. Smedley, ``Visual programming for robot control,''
  \emph{IEEE Symposium on Visual Languages.}, 1998.

\bibitem{ref09}
S.~H. Kim and J.~W. Jeon, ``Programming lego mindstorms nxt with visual
  programming,'' \emph{ICCAS}, 2007.

\bibitem{ref10}
C.~Datta, C.~Jayawardena, I.~H. Kuo, and B.~A. MacDonald, ``Robostudio: A
  visual programming environment for rapid authoring and customization of
  complex services on a personal service robot,'' \emph{IROS}, 2012.

\bibitem{ref11}
J.~Trower and J.~Gray, ``Blockly language creation and applications: Visual
  programming for media computation and bluetooth robotics control,'' \emph{ACM
  Technical Symposium on Computer Science Education.}, 2015.

\bibitem{ref12}
B.~D. Argall, S.~Chernova, M.~Veloso, and B.~Browning, ``A survey of robot
  learning from demonstration,'' \emph{RAS}, 2009.

\bibitem{ref13}
J.~Lee, ``A survey of robot learning from demonstrations for human-robot
  collaboration,'' \emph{arXiv preprint arXiv:1710.08789}, 2017.

\bibitem{ref14}
A.~Billard, S.~Calinon, R.~Dillmann, and S.~Schaal, ``Robot programming by
  demonstration,'' \emph{Springer Handbook of Robotics.}, 2008.

\bibitem{ref15}
H.~Veeraraghavan and M.~Veloso, ``Teaching sequential tasks with repetition
  through demonstration,'' \emph{International Foundation for Autonomous Agents
  and Multiagent Systems}, 2008.

\bibitem{ref16}
M.~Hersch, F.~Guenter, S.~Calinon, and A.~Billard, ``Dynamical system
  modulation for robot learning via kinesthetic demonstrations,''
  \emph{Transactions on Robotics}, 2008.

\bibitem{ref17}
C.~Kohrt, R.~Stamp, A.~Pipe, J.~Kiely, and G.~Schiedermeier, ``An online robot
  trajectory planning and programming support system for industrial use,''
  \emph{Robotics and Computer-Integrated Manufacturing}, 2013.

\bibitem{ref19}
M.~Kyrarini, M.~A. Haseeb, D.~Ristic-Durrant, and A.~G. aser, ``Robot learning
  of object manipulation task actions from human demonstrations,'' \emph{Facta
  Universitatis, Series: Mechanical Engineering}, 2017.

\bibitem{ref18}
A.~Mohseni-Kabir, C.~Rich, S.~Chernova, C.~L. Sidner, and D.~Miller,
  ``Interactive hierarchical task learning from a single demonstration,''
  \emph{International Conference on Human-Robot Interaction.}, 2009.

\bibitem{ref20}
C.~Finn, T.~Yu, T.~Zhang, P.~Abbeel, and S.~Levine, ``Oneshot visual imitation
  learning via meta-learning,'' \emph{arXiv preprint arXiv:1709.04905}, 2017.

\bibitem{dasari2019robonet}
S.~Dasari, F.~Ebert, S.~Tian, S.~Nair, B.~Bucher, K.~Schmeckpeper, S.~Singh,
  S.~Levine, and C.~Finn, ``Robonet: Large-scale multi-robot learning,''
  \emph{arXiv:1910.11215}, 2019.

\bibitem{ref23}
O.~Kilinc, Y.~Hu, and G.~Montana, ``Reinforcement learning for robotic
  manipulation using simulated locomotion demonstrations,'' \emph{arXiv
  preprint arXiv:1910.07294}, 2019.

\bibitem{ref24}
S.~Gu, E.~Holly, T.~P. Lillicrap, and S.~Levine, ``Deep reinforcement learning
  for robotic manipulation,'' \emph{CoRR}, 2016.

\bibitem{sasaki2017study}
H.~Sasaki, T.~Horiuchi, and S.~Kato, ``A study on vision-based mobile robot
  learning by deep q-network,'' in \emph{Conference of the Society of
  Instrument and Control Engineers of Japan (SICE)}.\hskip 1em plus 0.5em minus
  0.4em\relax IEEE, 2017.

\bibitem{wiki_learning}
\BIBentryALTinterwordspacing
\emph{Learning}. [Online]. Available:
  \url{https://en.wikipedia.org/wiki/Learning}
\BIBentrySTDinterwordspacing

\bibitem{h-mem}
\BIBentryALTinterwordspacing
\emph{Human Memory}. [Online]. Available: \url{http://www.human-memory.net}
\BIBentrySTDinterwordspacing

\bibitem{ref26}
D.~Vernon, M.~Beetz, and G.~Sandini, ``Prospection in cognition: The case for
  joint episodic-procedural memory in cognitive robotics,'' \emph{Frontiers in
  Robotics and AI}, 2015.

\bibitem{ref25}
G.~Sarthou, A.~Clodic, and R.~Alami, ``Ontologenius : A long-term semantic
  memory for robotic agents,'' \emph{IEEE International Conference on Robot \&
  Human Interactive Communication (IEEE Ro-MAN)}, 2019.

\bibitem{ref27}
M.~Sukhwani, V.~Duggal, and S.~Zahrai, ``Dynamic knowledge graphs as semantic
  memory model for industrial robots,'' \emph{arXiv:2101.01099}, 2021.

\bibitem{ref29}
D.~Stachowicz and G.~M. Kruijff, ``Episodic-like memory for cognitive robots,''
  \emph{IEEE Transactions on Autonomous Mental Development}, 2012.

\bibitem{ref30}
J.~Rothfuss, F.~Ferreira, E.~E. Aksoy, Y.~Zhou, and T.~Asfour, ``Deep episodic
  memory: Encoding, recalling, and predicting episodic experiences for robot
  action execution,'' \emph{CoRR}, 2018.

\bibitem{ref31}
E.~Castro and R.~Gudwin, ``An episodic memory for a simulated autonomous
  robot,'' \emph{Robocontrol}, 2010.

\bibitem{ref21}
T.~M. Mitchell, R.~M. Keller, and S.~T. Kedar-Cabelli, ``Explanation-based
  generalization: A unifying view,'' \emph{Machine learning}, 1986.

\bibitem{ref22}
R.~Dillmann, ``Teaching and learning of robot tasks via observation of human
  performance,'' \emph{Robotics and Autonomous Systems}, 2004.

\bibitem{mandal}
A.~{Mandal}, D.~{Sharma}, M.~{Sukhwani}, R.~{Jetley}, and S.~{Sarkar},
  ``Improving safety in collaborative robot tasks,'' in \emph{INDIN}, 2019.

\bibitem{thining}
\BIBentryALTinterwordspacing
\emph{Thinking Robots}. [Online]. Available: \url{https://thinkingrobots.ai}
\BIBentrySTDinterwordspacing

\bibitem{rinkus2004neural}
G.~J. Rinkus, ``A neural model of episodic and semantic spatiotemporal
  memory,'' in \emph{COGSCI}, 2004.

\bibitem{nuxoll2012enhancing}
A.~M. Nuxoll and J.~E. Laird, ``Enhancing intelligent agents with episodic
  memory,'' \emph{Cognitive Systems Research}, 2012.

\bibitem{starzyk2009spatio}
J.~A. Starzyk and H.~He, ``Spatio--temporal memories for machine learning: A
  long-term memory organization,'' \emph{Transactions on Neural Networks},
  2009.

\bibitem{lee2018hierarchical}
W.-H. Lee and J.-H. Kim, ``Hierarchical emotional episodic memory for social
  human robot collaboration,'' \emph{Autonomous Robots}, 2018.

\bibitem{wang2012neural}
W.~Wang, B.~Subagdja, A.-H. Tan, and J.~A. Starzyk, ``Neural modeling of
  episodic memory: Encoding, retrieval, and forgetting,'' \emph{Transactions on
  neural networks and learning systems}, 2012.

\bibitem{subagdja2015neural}
B.~Subagdja and A.-H. Tan, ``Neural modeling of sequential inferences and
  learning over episodic memory,'' \emph{Neurocomputing}, 2015.

\end{thebibliography}

\end{document}